\documentclass[review]{imag-ms-template}

\usepackage{amsmath}
\usepackage{bm}
\usepackage{adjustbox}

\usepackage{caption}
\usepackage{subcaption}
\usepackage{multirow}

\usepackage{todonotes}

\RestyleAlgo{ruled}
\DeclarePairedDelimiter\abs{\lvert}{\rvert}%
\DeclarePairedDelimiter\norm{\lVert}{\rVert}%

\let\cite\parencite

\makeatletter
\let\oldabs\abs
\def\abs{\@ifstar{\oldabs}{\oldabs*}}
\let\oldnorm\norm
\def\norm{\@ifstar{\oldnorm}{\oldnorm*}}
\makeatother

\SetKwComment{Comment}{/* }{ */}

\title{MUSTER: Longitudinal Deformable Registration by Composition of Consecutive Deformations}
\author{Edvard O. S. Grødem,$^{1,2\ast}$ Donatas Sederevičius,$^{1}$ Esten H. Leonardsen,$^{2,3}$ \\ Bradley J. MacIntosh,$^{1,4}$ Atle Bjørnerud,$^{1}$ Till Schellhorn,$^{1}$ Øystein Sørensen,$^{2}$\\ Inge Amlien,$^{2}$ Pablo F. Garrido,$^{2}$ Anders M. Fjell,$^{2}$ \\
{\small $^{1}$Computational Radiology \& Artificial Intelligence unit, Division of Radiology and Nuclear Medicine,}\\
{\small Oslo University Hospital, Oslo, Norway}\\
{\small $^{2}$Center for Lifespan Changes in Brain and Cognition, Department of Psychology,}\\
{\small University of Oslo, Oslo, Norway}\\
{\small $^{3}$Section for Precision Psychiatry, Oslo University Hospital \& Institute of Clinical Medicine,}\\
{\small University of Oslo, Oslo, Norway}\\
{\small $^{4}$Department of Medical Biophysics, Sunnybrook Research Institute,}\\ 
{\small University of Toronto, Toronto, Canada}\\
{\small $^\ast$Correspondence: edvardgr@uio.no}
}

\begin{document}
\maketitle

\keywords{Registration, Diffeomorphism and Longitudinal Analysis}

\begin{abstract}
Longitudinal imaging allows for the study of structural changes over time. One approach to detecting such changes is by non-linear image registration. This study introduces Multi-Session Temporal Registration (MUSTER), a novel method that facilitates longitudinal analysis of changes in extended series of medical images. MUSTER improves upon conventional pairwise registration by incorporating more than two imaging sessions to recover longitudinal deformations. Longitudinal analysis at a voxel-level is challenging due to effects of a changing image contrast as well as instrumental and environmental sources of bias between sessions. We show that local normalized cross-correlation as an image similarity metric leads to biased results and propose a robust alternative. We test the performance of MUSTER on a synthetic multi-site, multi-session neuroimaging dataset and show that, in various scenarios, using MUSTER significantly enhances the estimated deformations relative to pairwise registration. Additionally, we apply MUSTER on a sample of older adults from the Alzheimer's Disease Neuroimaging Initiative (ADNI) study. The results show that MUSTER can effectively identify patterns of neuro-degeneration from T1-weighted images and that these changes correlate with changes in cognition, matching the performance of state of the art segmentation methods. By leveraging GPU acceleration, MUSTER efficiently handles large datasets, making it feasible also in situations with limited computational resources.
\end{abstract}

\section{Introduction}
Registration of medical images is the process of finding a spatial transformation, either linear or non-linear, such that the images are aligned. Medical image registration is a crucial operation, widely used for aligning images to a common template \cite{fonov2011unbiased} or to align different image modalities of the same subject \cite{maes1997multimodality}. Longitudinal analysis of a subject can be performed using non-linear registration, for instance for cancer tracking \cite{clatz2005robust, fuster2022quantification} or the study of neurodegeneration in the case of dementia \cite{holland2011nonlinear, avants2008symmetric}. 

There are many popular methods for non-linear registration of medical images, mostly based on pairwise registration of an image to a template and some that are applied to longitudinal analysis of longer series of images.  

"Symmetric Image Normalization Method" (SyN) \cite{avants2008symmetric} deforms two images to the middle time point between the images using two non-linear deformation fields. An optimization procedure is performed where the deformations are updated using a gradient step and the inverse of the deformations is found with a fixed point algorithm. SyN is implemented in the toolbox Advanced Normalization Tools (ANTs) \cite{avants2009advanced, tustison2021antsx}. 

"Diffeomorphic Anatomical Registration using Exponentiated Lie Algebra" (DARTEL) \cite{ashburner2007fast} does pairwise registration by integrating a constant flow field using the Log-Euclidean framework. The method is described in Section \ref{se:log_euc} and our approach to longitudinal analysis generalizes this method to more than two images.

The method "Large Deformations Diffeomorphic Metric Mapping" (LDDMM) \cite{beg2005computing} and other similar methods are often referred to as geodesic shooting. In these methods a Riemannian metric \cite{lee2018introduction} is defined on the manifolds of diffeomorphic deformations, and an initial vector momentum field is optimized such that a geodesic path between two images is found. 

Building on LDDMM there are a group of techniques for doing longitudinal analysis of more than two images, named geodesic regression. \textcite{hong2012simple} presents a simple geodesic regression method by doing pairwise LDDMM registration between the first image and all subsequent images. A longitudinal regression of the deformation between the first image to all other image is calculated by a weighted average of the initial vector momentums.  

Further work on geodesic regression is presented in \textcite{hinkle2012polynomial, singh2013vector, thomas2013geodesic, singh2015splines}, where a template as well as a geodesic path from the template to the images are optimized.
These methods are attractive since they guarantee a diffeomorphic deformation and due to their geodesic nature, also find the shortest paths on the Riemannian manifold defined by the image similarity metric. They are however quite computationally expensive \cite{sotiras2013deformable, jena2024deep} and therefore most of these methods are applied to surfaces or 2D images. The methods also require a Riemannian metric, which exclude some similarity metrics such as mutual information and local normalized cross-correlation which limit their applications. 

\textcite{ashburner2013symmetric} introduced a framework based on LDDMM that included bias correction, and both rigid and non-linear registration. In their framework, all images in a time series are deformed to a subject template. Their work differs from ours in that all images are independently deformed to the template, while our work take the composability of the deformations into consideration. 

\textcite{ding2019fast} used two U-nets \cite{ronneberger2015u} in series to predict the initial momentum of a geodesic shooting based registration. They extend the method to longitudinal series by the simple geodesic regression method described by \textcite{hong2012simple}. 

For a comprehensive review of registration methods as well as image similarity metrics see \textcite{sotiras2013deformable}.

In the current work, we present an algorithm to register a time series of possibly more than two images to each other. We call our algorithm: MUSTER - Multi Session Temporal Registration. 

There are two core innovations to MUSTER:

\begin{itemize}
    \item MUSTER does both linear and non-linear registration of all images to all other images in a time series by composing consecutive deformations.
    \item MUSTER uses a modified version of local normalized cross-correlation which gives a less biased estimate of the deformation fields.
\end{itemize}

Image registration is an ill-posed problem \cite{sotiras2013deformable}. Adding constraints and regularization is therefore essential for giving consistent outputs. By considering that a deformation from time point 1 to time point 3 has to go through time point 2, MUSTER adds an extra constraint to the registration process compared to pairwise registration from time point 1 to 3. This restricts and guide the estimated deformation to follow the true trajectory. Medical imaging contains artifacts such as noise, change in contrast, and bias fields. By considering all images in a time series, we can estimate these artifacts which also increase the robustness of MUSTER compared to pairwise options. 

\begin{figure}
    \centering
    \includegraphics[width=0.6\linewidth]{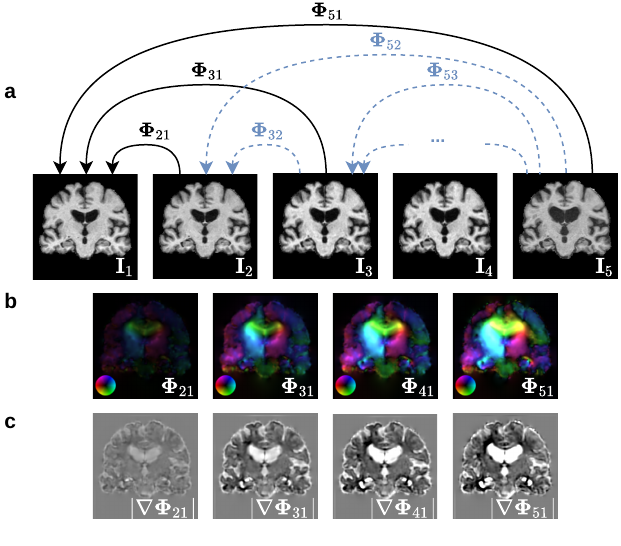}
    \caption{\textbf{a}: MUSTER does deformable registration between all images of a time series. Here a series of T1-weighted brain scans from ADNI is displayed. The arrows indicates the deformations relating the images of the time series. For simplicity most of the deformations are not shown. \textbf{b}: The deformation fields that deform the images to the first image $I_1$ are visualized as colormaps. The color hue shows the deformation direction orthogonal to the image plane, and the color intensity shows the magnitude of the deformation. \textbf{c}: The determinant of the spatial Jacobian of the deformation field. Dark regions show contraction, and bright regions show expansion of the tissue.}
    \label{fig:overview}
\end{figure}

The method is implemented to utilize graphics processing units (GPUs) to accelerate the registration model. This makes the algorithm run fast, and the deformation between 12 3D images of the brain can be estimated within 3 minutes. MUSTER is therefore a viable tool for processing large datasets without the need for large compute nodes. Fig. \ref{fig:overview} illustrates a subject from ADNI processed with MUSTER.   

In this paper we first present the background with terminology and mathematical concepts. We then present the proposed method for longitudinal registration. Furthermore, we highlight the issue of using local normalized cross-correlation as an image similarity metric and suggest an alternative loss function. We verify our method by comparing it to other approaches on two experimental setups. First we test how well MUSTER can estimate deformations from a synthetic dataset of longitudinal brain scans. Second we use MUSTER to do analysis on a subset of the ADNI dataset, and relating the deformations to change in cognitive scores. 

\section{Method}

\subsection{Background on Longitudinal Deformations} 
Intra-subject registration aims to recover the movement of anatomical structure over time. Let $\bm A_{1} : \bm \Omega \to \mathbb{R}^3$ represent the anatomical structure, or tissue, at reference time $t_1$. The spatial domain $\bm \Omega \subset \mathbb{R}^3$ represents the space which the anatomical structure occupies. The anatomical structure that changes over time can then be expressed as the $\bm A_i(\bm x) = \bm A_i(\bm \Phi_{1i}(\bm x))$, where $\bm \Phi_{1i}(\bm x): \bm \Omega \to \bm \Omega$ is a deformation map from $t_1$ to $t_i$.

We assume that all changes in the $ \bm A $ are solely due to the tissue expanding, contracting, or being displaced. This assumption implies that the movement of $ \bm{A}_t $ is described by a diffeomorphic mapping between all recorded time points. 

A transformation is said to be a diffeomorphic mapping if it has the following properties: 
\begin{itemize}
    \item The transformation is continuously differentiable.
    \item The inverse of the transformation exists.
    \item The inverse of the transformation is continuously differentiable.
\end{itemize}

These properties ensure that anatomical tissue retains its physical integrity—specifically, it cannot pass through itself and must maintain a positive volume.

This is a common and often desired assumption in many non-linear registration methods. Note, however, that tissue can also change in ways not modeled by diffeomorphic transformations, such as changes in intensity (e.g., due to contrast uptake), the appearance of new structures, or the removal of tissue (e.g., due to surgery). Our method is somewhat robust to changes in intensity due to the selection of similarity metric, however it does not handle objects appearing or disappearing, such as a tumor appearing between sessions, due to the assumption of diffeomorphic deformations. 

The deformation field $\bm{\Phi}_{ji}(\bm x) :\bm{\Omega} \to \bm{\Omega} $ maps the anatomical structure at time $t_j$ to that at time $t_i$. Applying this deformation to $\bm A_{j}(\bm x )$, we obtain $\bm{A}_{j}(\bm{\Phi}_{ji}(\bm x)):=  \bm{A}_{j} \circ \bm{\Phi}_{ji} (\bm x) = \bm{A}_{i}(\bm x) $, aligning the anatomical structures at the two time points.


An image $\bm I_i$ is an observation of $ \bm{A}_i(\bm x)$.  The image can be seen as a function $\bm I_i : \Omega_{\mathbb{I}} \to \mathbb{R}$, which maps $\bm A_i$ on discrete regular grid $\bm \Omega_\mathbb{I}$ to intensities. When doing longitudinal registration we do so by imaging at $ N $ distinct time points, $ \{t_i\}_{i=1}^N $. This results in a series of images $ \{\bm I_i\}_{i=1}^N $.

 Given the deformation $\bm \Phi_{ji}$ and the image $ \bm{I}_j $, we can approximate $ \bm{I}_i $ by interpolating $ \bm{I}_j $ on an irregular grid given by $ \bm{\Phi}_{ji}^i =  \{\bm{\Phi}_{ji}(\bm x) \; \forall \; \bm x \in \bm \Omega_{\mathbb{I}})\}$. $\bm{\Phi}_{ji}^i$ is the discrete version of $\bm{\Phi}_{ji}$ where the superscript "i" indicates that the function has been evaluated on the regular imaging grid at time $t_i$. The operation of deforming $\bm I_j$ to $\bm I_i$ is denoted 
 \begin{equation}
     \bm I_j \circ \bm \Phi_{ji}^i \approx \bm I_i.
 \end{equation}
 Even with a perfect deformation, both due to image artifacts and interpolation errors, $\bm \Phi_{ji}^i$ will only partially correctly transform one image into another. One can also deform $\bm I_i$ to approximate $\bm I_j$ by using the inverse deformation $\bm \Phi_{ij} = \bm \Phi_{ji}^{-1}$: 
 \begin{equation}
     \bm I_i \circ (\bm \Phi_{ji}^i)^{-1} = \bm I_i \circ \bm \Phi_{ij}^j \approx \bm I_j.
 \end{equation}

A diffeomorphic transformation on a discrete grid can be constructed by numerically integrating an ordinary differential equation (ODE). For each grid point $\bm x$, we simulate the trajectory of a particle moving in a time-dependent flow (velocity) field $\bm \phi(\bm x,t)$: 
\begin{subequations}\label{eq:ode}
    \begin{equation}
        \frac{d\bm \Phi^i(\bm x, t)}{dt} = \bm{\phi}(\bm \Phi^i(\bm x, t), t),
    \end{equation}
    \begin{equation}
        \bm \Phi^i(\bm x, t_i) = \bm x.
    \end{equation}
\end{subequations}

Again, the superscript "i" indicates for what time the discrete deformation field passed through the regular grid $\Omega_{\mathbb{I}}$ at time $t_i$. Fig. \ref{fig:deforms}a shows an illustration of the relationships between the deformation fields and their sub- and superscripts.

\subsection{MUSTER overview}
The key innovation of MUSTER is to construct all deformations between all images from the deformations between consecutive time points. To find the deformations that best describe the change in anatomical tissue, we deform each image to all other images and calculate an image similarity metric. Then, a gradient-based optimizer based on auto-differentiation \cite{paszke2017automatic} is used to update the deformation fields. In addition, regularization losses are added to ensure that the deformation is physically feasible. This gives the loss function
\begin{equation}
    L =  \sum_{i \in [1, N]} \sum_{j \in [1, N] \setminus i} \text{Sim}(\bm I_i, \bm I_j \circ \Phi_{ji}^i) + L_{\text{reg}} ,
\end{equation}
where $\text{Sim}(\cdot, \cdot)$ is an image similarity metric (e.g. L2 norm, local normalized cross-correlation) and $L_{\text{reg}}$ is a regularization term that enforces smoothness and physical plausibility of the deformation fields.

\begin{figure}
    \centering
    \includegraphics[width=0.6\textwidth]{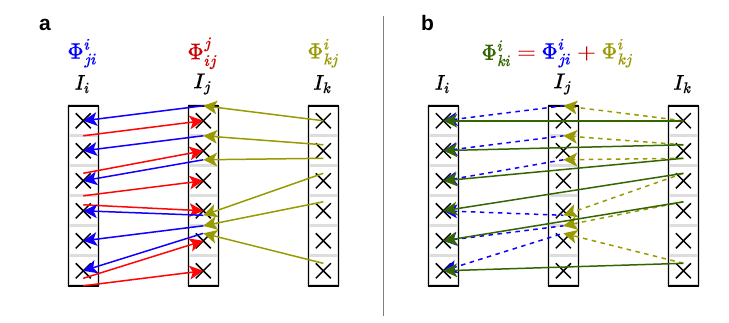}
    \caption{Illustration of the deformation fields on 1-dimentional images. The rectangles represent voxels, the $\times$ represents the centers of voxels and the arrows are deformations. $\bm \Phi_{ji}^i$ is the deformation that "pulls" $\bm I_j $ to $\bm I_i$ and $\bm \Phi_{ij}^j \approx (\bm \Phi_{ji}^i)^{-1}$ is the deformation that "pulls" $\bm I_i$ to $\bm I_j$. $\bm \Phi_{kj}^i$ is the deformation of the voxels from $\bm I_i$ in the interval between $\bm I_k$ and $\bm I_j$. \textbf{a} illustrates the meaning of the sub- and superscripts. \textbf{b} shows how two consecutive deformations can be combined to construct a combined deformation.}
    \label{fig:deforms}
\end{figure}

Given that we have the two discrete deformations $\bm \Phi_{kj}^j$ and $\bm \Phi_{ji}^i$, one can approximate the composite deformation $\bm \Phi_{ki}^i$ using some interpolation method. In our method, we used trilinear interpolation to approximate $\bm \Phi_{ij} \circ \bm \Phi_{jk} $ with $ \bm \Phi_{ij}^j \circ \bm \Phi_{jk}^k $ where
\begin{equation} \label{eq:cum_deformation}
    \bm \Phi_{ki}^i = \bm \Phi_{ji}^i + (\bm \Phi_{kj}^i - \Omega_{\mathbb I}) \approx  \bm \Phi_{kj}^j \circ \bm \Phi_{ji}^i =  \bm \Phi_{ji}^i + (\bm \Phi_{kj}^j - \Omega_{\mathbb I}) \circ \bm \Phi_{ji}^i.
\end{equation}
Fig. \ref{fig:deforms}b illustrates how two discrete deformation fields can be combined using interpolation to create a composed deformation field. 

\begin{figure}
    \centering
    \includegraphics[width=0.5\linewidth]{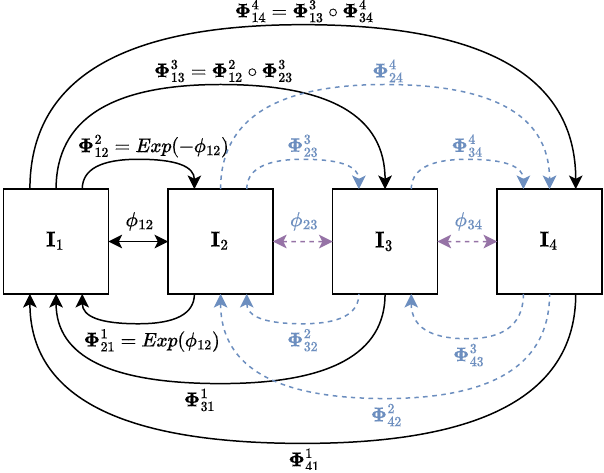}
    \caption{Overview of the deformations relating the images of a image series. $\phi_{ij}$ denotes the flow field between the consecutive images $\bm I_i$ and $\bm I_j$. $\bm \phi_{ij}^j$ denotes deformation on a grid between $\bm I_i$ and $\bm I_j$. All deformations between all images can be constructed by composing the consecutive deformations. }
    \label{fig:enter-label}
\end{figure}

To do longitudinal registration of $N$ time points, we therefore only need to find $(N-1)$ diffeomorphic deformations between the consecutive sessions in the forward and backward directions. The rest of the $N(N-2)$ deformations between the sessions can be calculated from these consecutive deformations using Eq. \ref{eq:cum_deformation}.

\subsection{Parameterization of consecutive diffeomorphism}
\label{se:log_euc}
 The consecutive deformations must be parameterized such that they are diffeomorphic mappings. Registration to multiple images also makes the need for an efficient parameterization essential to ensure that the computational and memory requirements of the algorithm are within practical limits. 
 
A simple method to obtain a diffeomorphic transformation is by integrating a stationary deformation flow $\bm \phi_{ij}$  with an initial position on a regular grid \cite{arsigny2006log}. While Euler integration is an option for this purpose, it often requires a high number of time steps for accurate results. In line with existing solutions such as \textcite{ashburner2007fast, balakrishnan2019voxelmorph, iglesias2023easyreg}, we employ the Log-Euclidean framework \cite{arsigny2006log} for parameterization of diffeomorphic deformations. The Log-Euclidean framework offers an efficient alternative to Euler integration through the "scaling-and-squaring" technique. This method accomplishes a diffeomorphic transformation in $log_2(n)$ iterations, as opposed to the $n$ iterations needed with Euler integration, in order to obtain a similar accuracy.  

The Log-Euclidean framework utilizes the theory of Lie groups and therefore uses the terms logarithm and exponent. The deformation flow is called the logarithm of the deformation field and vice versa the deformation field is called the exponential of the deformation flow $\bm \Phi = Exp(\bm \phi)$. The computation of $Exp(\bm \phi)$ is given in Algorithm \ref{alg:exp}.  

\begin{algorithm}[h] 
\caption{$Exp(\bm\phi)$}\label{alg:exp}
\KwData{$\bm\phi$, $N_{iter}$}
\KwResult{$\bm \Phi$}
$\bm \Phi \gets \bm \Omega_{\mathbf{I}} +  \frac{1}{2^{N_{iter}}}\bm\phi$;\\
\For{$i \gets 0...N_{iter}$}{
  $\bm \Phi \gets \bm \Phi \circ \bm \Phi$; \\
  \Comment{Due to boundary effects, it is better instead to compute \\
            $\bm \Phi \gets \bm \Phi + (\bm \Phi - \bm \Omega_{\mathbb{I}})\circ \bm \Phi$;}
}
\Return $\bm \Phi$;
\end{algorithm}

Using the Log-Euclidean framework has two important features. As mentioned above, it produces diffeomorphic transformations if the deformation flow is smooth. Second, the inverse of the deformation is easily computed as $\bm \Phi^{-1} = Exp(\bm \phi)^{-1} = Exp(- \bm \phi)$. An intuitive explanation of the Log-Euclician framework is given by \textcite{ashburner2007fast}.

 In our method we parameterize the consecutive deformations with $N-1$ stationary deformation flows $ \bm \phi = \{\bm \phi_{i,i+1}, i \in \{1, N-1\} \}$. The deformations between the consecutive time points are calculated as follows:
\begin{subequations}
    \begin{equation}
        \bm \Phi_{i+1,i}^{i} = Exp( \bm \phi_{i,i+1}) \quad  \forall \quad  i \in \{1, N-1\},
    \end{equation}
    \begin{equation}
        \bm \Phi_{i,i+1}^{i+1} = (\bm \Phi_{i+1,i}^{i})^{-1} = Exp(-\bm \phi_{i,i+1}) \quad  \forall \quad  i \in \{1, N-1\}.
    \end{equation}
\end{subequations}

\subsection{Regularization}
In image registration, there are often many possible deformations that can explain the changes observed between images. Therefore, it is necessary to regularize the deformation flows to obtain plausible and physically meaningful solutions. As previously mentioned, ensuring that the deformations are diffeomorphic mappings is one way to achieve this. However, the Log-Euclidean framework will only produce diffeomorphic deformations if the stationary vector field is sufficiently smooth \cite{arsigny2006log}. Simply optimizing the parameters of a discrete deformation flow may not necessarily yield a diffeomorphic mapping. 

Smoothness of the deformation flow can be enforced either by adding a regularization term to the loss function or through explicit parameterization of the flow field. In MUSTER the flow field is smoothed with a Gaussian kernel, implemented using Fast Fourier Transforms (see Appendix \ref{se:gaussian_smoothing} for implementation details). This ensures a minimum level of smoothness through patameterization. Additionally, we use a regularizing loss that encourages spatial smoothness by applying the Frobenius norm to the spatial Jacobian of the deformation flow. This forces the deformation flow towards being $C^1$ continuous, equivalent to having a smoothness prior on the deformation flow. The loss is given by
\begin{equation} \label{eq:reg_ss}
    L_{ss} = \frac{1}{N-1}\sum_{i=1}^{N-1}\norm{ \bm \nabla \bm \phi_{i,i+1} }_F^2,
\end{equation}
where $\norm{ \cdot }_F$ denotes the Frobenius norm given by
\begin{equation}
    \norm{\bm A}_F = \sqrt{\sum_{k=1}^{N_A} \sum_{l=1}^{M_A} \abs{a_{kl}}^2},
\end{equation}
where $\bm A$ is an $N_A$ by $M_A$ matrix and $a_{kl}$ is the indexed element in the matrix. The Jacobian $\nabla \bm \phi_{i,i+1}$ is calculated using finite difference methods at the resolution of the deformation flow grid.

For stability of the optimization process and for data with very small deformations we added an L2 loss \cite{ashburner2013symmetric}
\begin{equation} \label{eq:reg_ss}
    L_{L2} = \frac{1}{N-1}\sum_{i=1}^{N-1}\norm{ \bm \phi_{i,i+1} }_2^2
\end{equation}
on the flow field. This is equivalent to putting a Gaussian prior on the magnitude over the flow field. 

In some applications, such as modeling brain changes over time, we expect the deformations to be smooth in time. That is, when observing changes in anatomy from $t_i$ to $t_j$, similar changes are often observed from $t_j$to $t_k$. Assuming that the deformation flow is proportional to time, we define the following temporal smoothness regularization loss:
\begin{equation} \label{eq:reg_ts}
    L_{ts} = \frac{1}{N-2}\sum_{i=2}^{N-1} \norm{ \frac{\bm \phi_{i,i-1}}{t_{i}-t_{i-1}} - \frac{\bm \phi_{i,i+1}}{t_{i+1}-t_{i}} }_F^2 .
\end{equation}

The total regularizing loss is then given by 
\begin{equation} \label{eq:reg_sum}
    L_{reg} = \alpha_{ss} L_{ss} + \alpha_{L2} L_{L2} + \alpha_{ts} L_{ts}
\end{equation}
where $\alpha_{ss}, \; \alpha_{L2}, \;\alpha_{ts}$ are hyper parameters controlling the strength of the spatial smoothness, deformation magnitude and temporal smoothness regularization, respectively.

\subsection{Rigid Registration}
Our method assumes that the images in the series are roughly aligned with one another using either rigid or affine registration. For large deformation in the tissue, it can be hard to find the linear transformation that ensures that the non-linear deformations are as small as possible. We therefore do rigid registration in parallel with the deformable registration described above. We do so by parameterizing a rigid adjustment in terms of Euler angles and a translation. All images except the first image are linearly adjusted inn additional to the deformation field using the total deformation: 
\begin{equation}
    \bm \Phi_{ij \text{Total}}^j = \bm \Phi_{ij}^j + (\bm \Phi_{j \text{Lin}}^j - \bm \Phi_\mathbf{I}) .
\end{equation}

\subsection{Image Similarity Metric for Longitudinal Registration}
\label{se:loss_func}
Most medical imaging methods do not provide a quantitative mapping between underlying tissue properties and image intensities. The image pixel-wise intensities, in addition to underlying tissue properties, depend on multiple scanner specific factors that may vary between scanning sessions. Image noise and global artifacts may also vary between sessions.  These factors make similarity metrics like mean squared error (MSE) unsuitable for accurately estimating the alignment of anatomical structures. 

A popular similarity metric is the local normalized cross-correlation (LNCC) used by \textcite{avants2008symmetric, balakrishnan2019voxelmorph} which is known to be invariant to local changes in contrast. 
The LNCC is defined as the average of the normalized cross-correlation in a square window $R$ that is shifted over the two images $\bm I_i, \bm I_j$. For consistency with the other loss functions in this paper we use $1 -$ "the conventional LNCC" such that the minimizing LNCC is the same as maximizing the correlation. For each region $R$ the cross-correlation is computed as 
\begin{equation}
    \label{eq:LNCC}
    \text{LNCC}_R = 1 - \frac{1}{|R|} \frac{\sum_{\bm{x} \in R} \bar{\bm I}_{i\bm{x}} \bar{\bm I}_{j\bm{x}}}{\sqrt{S_{i}^2 S_{j}^2}}
\end{equation}
where $\bar{\bm I}_{i\bm{x}} = \bm I_{i\bm{x}} - 1/|R|\sum_{\bm{x}' \in R}\bm I_{i\bm{x}'}$ and $S_{i}^2 = 1/|R|\sum_{\bm{x}' \in R} (\bar{\bm I}_{i\bm{x}'R})^2$ is the maximum likelihood estimate of the variance of $\bm I_i$ in $R$. $|R|$ denotes the number of voxels in $R$. The total loss over the image is then computed as 
\begin{equation}
\label{eq:sum_lncc}
    \text{LNCC} = \frac{1}{N_R}\sum_{R\in \bm I} \text{LNCC}_R
\end{equation}
Where $N_R$ is the number of regions in the overlapping images. 

To see why LNCC might fail we set up a simple model for the image intensities. Assuming that for a small enough region $R$ there is a linear relationship between the intensities of the observed images and the true image. Additionally, we assume image noise to be Gaussian.  Within one region this gives the following model:
\begin{subequations}
\label{eq:model_images}
\begin{equation}
    \bm I_{i\bm{x}} = a_{iR} \bm I_{\bm{x}} + b_{iR} + \bm \epsilon_{i\bm{x}} ,
\end{equation}
\begin{equation}
    \bm I_{j\bm{x}} = a_{jR} \bm I_{\bm{x}} + b_{jR} + \bm \epsilon_{j\bm{x}},
\end{equation}
\begin{equation}
    \bm \epsilon_{i\bm{x}} \sim N(0, \sigma_i),
\end{equation}
\begin{equation}
    \bm \epsilon_{j\bm{x}} \sim N(0, \sigma_j),
\end{equation}
\end{subequations}
where $\sigma_i, \sigma_j$ are the global noise scales for each image. 

Since $\bm I_{i\bm{x}}$ and $\bm I_{j\bm{x}}$ are linear transformations of the true image $\bm I$, they can be related directly with a linear transformation: 
\begin{equation}
\label{eq:lin_relation}
\bm I_{j\bm{x}} = a_{R} \bm I_{i\bm{x}} + b_{R} + \bm \epsilon_{\bm x} 
\end{equation} 
where $a_{R}$ and $b_{R}$ are parameters relating the two images at in $R$ and $\bm \epsilon_{\bm x}$ is the combined normal distributed noise term with $\sigma^2= a_{R}^2  \sigma_i^2 +  \sigma_j^2$

Assuming the cross-correlation is calculated over a sufficiently large region (see Appendix \ref{apx:Expectation_lncc}), the expected loss for a region can then be expressed as:
\begin{equation}
    \mathbb{E}[\text{LNCC}_R] \approx 1- \frac{1}{\sqrt{1 + \frac{\sigma^2}{a_{R}^2 S_i^2}}} = 1 - \frac{1}{\sqrt{1 + \frac{1}{a_{R}^2\text{CNR}^2}}} \label{eq:E_lncc}
\end{equation}

where $S_i^2 = \frac{1}{N_R} \sum_R \bar{I}_{i\bm{x}}^2$ and CNR is the Contrast-to-Noise-Ratio given by $\text{CNR} = \frac{S_i}{\sigma}$. 

\begin{figure}[h]
    \centering
    \includegraphics[width=0.25\linewidth]{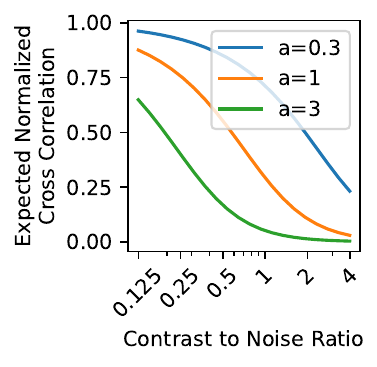}
    \caption{$E\left[\text{LNCC}_R\right]$ plotted as a function of CNR. }
    \label{fig:expected_ncc_loss_analytical}
\end{figure}

\begin{figure}
    \centering
    \includegraphics[width=0.75\linewidth]{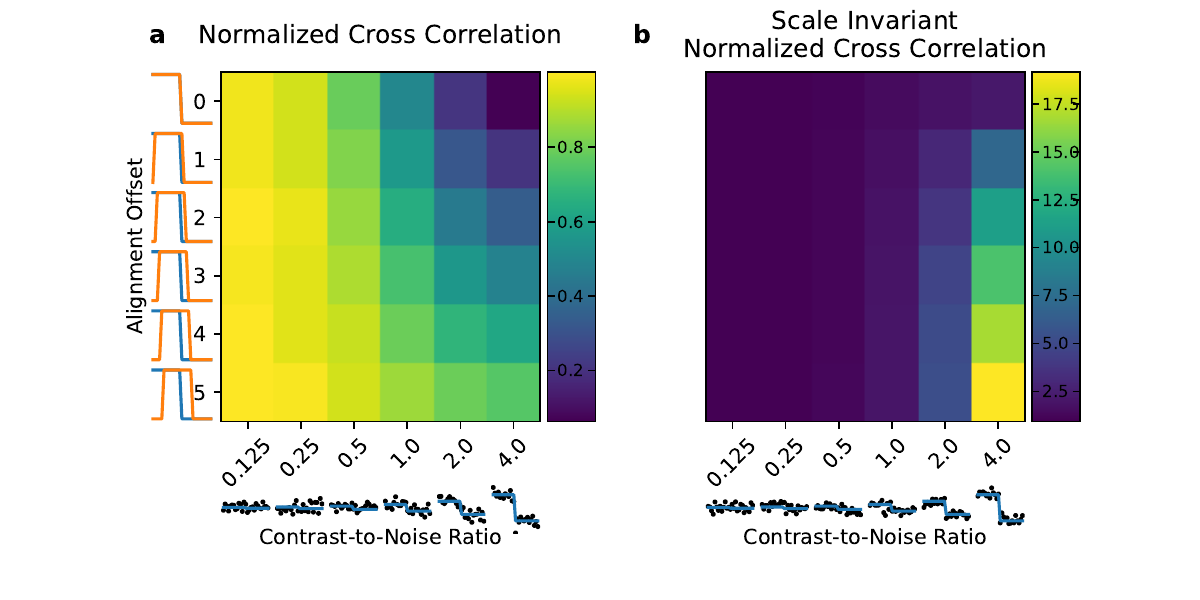}
    \caption{\textbf{a:} Local normalized cross-correlation as a function of contrast to noise ratio and alignment of two 1 dimensional images estimated using a Monte Carlo simulation. See Appendix \ref{se:cross_cor_exp} for implementation details.  The x-axis represent the CNR ratio and the y axis is the offset between the step functions. The offset and contrast to noise ratio is illustrated with a small plot for each axis. \textbf{b:} Scale invariant local normalized cross-correlation plotted for the Monte Carlo simulation as in \textbf{a}. }
    \label{fig:cross_correlation}
\end{figure}

From Eq. \ref{eq:E_lncc} and Fig. \ref{fig:expected_ncc_loss_analytical} we observe the following. The expected loss decreases with higher CNR of a region.  In regions with low CNR, such as white matter in T1-weighted Magnetic Resonance Images (MRI), cerebrospinal fluid (CSF), or air there are few features that can be used to estimate the deformation. In a probabilistic framework like ours, and in many other such as VoxelMorph \cite{balakrishnan2019voxelmorph, hoopes2022learning},  where priors are balanced against the data likelihood, LNCC reduce the influence of priors in low CNR regions. This results in less smooth deformation fields where there are few features to guide registration. Ideally, we should rely more heavily on priors in such regions to ensure smooth and physically plausible deformations. 

To address this issue, we employ a similarity metric derived from first principles that is similar to the loss function described in \textcite{pan2011recent}. However, this loss function is not used in medical registration and  \textcite{pan2011recent} did not point out the negative consequenses of using LNCC. 

We start by considering the model of Eq. \ref{eq:lin_relation}. The likelihood function of a region $R$ is given by:
\begin{equation}
    \text{NLL}_{R} \propto \sum_{\bm x\in R} \left(a_{R} \bm I_{i\bm x} + b_{R} - \bm I_{j\bm x}\right)^2.
\end{equation}
As  before we assume that $a_{R}$ and $b_{R}$ change slowly such that within $R$ they are approximately constant. We also assume that $\sigma_i \ll 1$. The maximum likelihood (ML) estimates for $a_{R}, b_{R}$ can then be derived analytically and are the same as in least square regression. Inserting for $a_{R}, b_{R}$ gives the following function:
\begin{equation}
    \text{NLL}_{R} \propto \text{SiLNCC}_R = \frac{1}{|R|} \left(\sum_{\bm x \in R}\bar{\bm I}_{j{\bm x}}^2 - \frac{\left(\sum_{\bm x \in  R}\bar{\bm I}_{i\bm x} \bar{\bm I}_{j\bm x} \right)^2}{\sum_{\bm x \in  R}\bar{\bm I}_{i\bm x}^2} \right).
\end{equation}
This loss looks similar to LNCC, but solves the normalization issue. We therefore call it scale invariant local normalized cross-correlation (SiLNCC).






In MUSTER the image similarity loss is calculated as the mean SiLNCC of all regions. Finding the deformation $\bm \Phi_{ij}^i$ that minimizes this is equivalent to finding the maximum likelihood estimation of the $\bm \Phi_{ij}^i$:  
\begin{equation} \label{eq:sim_loss}
    \text{Sim}(\bm I_i, \bm I_j \circ \bm{\Phi}_{ji}^i) = L_{\text{SiLNCC}}(\bm I_i, \bm I_j \circ \bm{\Phi}_{ji}^i) = \frac{1}{N_R}\sum_{R \in \Omega_\mathbb{I}} \text{SiLNCC}_R(\bm I_i, \bm I_j \circ \bm{\Phi}_{ji}^i).
\end{equation}

In Fig. \ref{fig:cross_correlation} the expected loss for LNCC and SiLNCC is plotted as a function of CTN and image alignment offset using a Monte Carlo estimation. We see that for a fixed offset the loss decreases for LNCC with higher CNR while the opposite is true for SiLNCC. This shows that SiLNCC is better suited to balance priors up against the likelihood of the data.      

\subsection{Implementation Details}
We carried out the implementation of MUSTER using PyTorch \cite{paszke2019pytorch}, taking advantage of its GPU acceleration and automatic differentiation capabilities. The optimization is performed using the Adam optimizer, coupled with a learning rate scheduler starting with a linear warmup for the first 20\% of iterations, followed by a cosine decay.

The image registration process was broken down into three stages. In each stage, the images were downsampled with a factor of $[4,2,1]$, and  deformation flows had a resolution of $[8,4,2]$ in relation to the full image resolution. At the beginning of each stage, we initialized the deformation flows by interpolating the results from the previous stage. The number of iterations set for each stage were $[200,200,100]$.

We used SiLNCC with a window size of 3 as the similarity metric. 

\subsection{Evaluation Metrics}
When the ground truth deformation is accessible we use several metrics to evaluate the performance of the deformation models. 

The Euclidean distance is a measure of the distance between two vector fields. The Euclidean distance can easily be calculated by 
\begin{equation}
    Eu(\bm \Phi^\star, \bm \Phi) = \frac{1}{|\bm  \Omega_{ROI}|}\sum_{\bm x\in \bm  \Omega_{ROI}} ||\bm \Phi^\star - \bm \Phi||_2
\end{equation}
where $\bm \Phi^\star$ is the ground-truth deformations, $\bm \Phi$ is the estimated deformation and $\bm \Omega_{ROI}$ is the set of points in the region of interest (ROI). 

The Pearson correlation coefficient (PCC) can be generalized for vectors, and is given by
\begin{equation}
    \text{PCC}(\bm \Phi^\star, \bm \Phi) = \frac{\sum_{\bm x \in \bm \Omega_{ROI}}\bar{\bm \Phi}^\star \cdot \bar{\bm \Phi} }{\sqrt{\sum_{\bm x \in \bm \Omega_{ROI}} \bar{\bm \Phi}^\star\cdot \bar{\bm \Phi}^\star} \sqrt{\sum_{\bm x \in \bm \Omega_{ROI}}\bar{\bm \Phi}\cdot \bar{\bm \Phi}}}
\end{equation}
where $\cdot$ is the per voxel dot-product and $\bar{\bm \Phi} = \bm \Phi -\frac{1}{|\bm \Omega_{ROI}|}\sum_{\bm \Omega_{ROI}} \bar{\bm \Phi} $. 

The Pearson Correlation Coefficient (PCC) quantifies how much information can be retrieved using a linear model. However, it does not provide insight into whether an algorithm systematically overestimates or underestimates the deformation. To assess the bias of the model, we use linear regression to relate the ground truth deformation to the estimated deformation: %
\begin{subequations}
    \begin{equation}
       \bm \Phi = \bm A \bm \Phi^\star+ \bm b,
    \end{equation}
    \begin{equation}
        B = \frac{1}{Dim(\bm A)}\text{Trace}(\bm A),
    \end{equation}
\end{subequations}
where $B$ represents the average of the diagonal elements of $\bm A$. If $\bm A$ is approximately a diagonal matrix, $B$ corresponds to the mean of its eigenvalues.

The value of $B$ serves as an indicator of bias. Specifically:
\begin{itemize}
    \item If $B$ is close to 1, the deformation field is unbiased.
    \item If $B > 1$, the deformation field is overestimated.
    \item If $B < 1$, the deformation field is underestimated.
\end{itemize}
    
\section{Experimental Validation}
\subsection{Simulated Deformations}
\label{se:syn_data}
 A synthetic dataset was generated to verify that our approach could estimate the deformation of tissue. The aim for this synthetic data was to mimic a real-world longitudinal imaging study like the ADNI dataset, where subjects are followed over a long period and where the images are acquired at different scanners. The images were, therefore, subject to scanner noise, contrast changes, bias fields and distortions. Between the sessions a synthetic deformation was applied to the images. The synthetic deformation represented a change in the brain due to atrophy, tumor growth or neurogenesis. 

The LCBC Traveling Brains dataset features MRI scans from 7 subjects, each scanned 18 to 22 times on 9 to 11 different scanners within a period of a month. We assumed that the tissue characteristics didn't change over this short period and that any intensity changes were due to imaging artifacts. 
From this dataset, we created 240 synthetic time series. Each series was constructed by randomly selecting a subject and then drawing 8 sessions from that subject's available data.

We generated a diffeomorphic deformation for each time series. Specifically, a continuous deformation flow was created for each subject, generated from white noise that was then smoothed using a Gaussian filter. The parameters for these filters were: 
\begin{itemize}
    \item Spatial frequency $\omega_{s}$: $[0.03, 0.1, 0.3] \frac{1}{\text{mm}}$
    \item Temporal frequency $\omega_{t}$: $3.0 \; \frac{1}{\text{mm}}$
\end{itemize}
Different spatial frequencies allow us to generate deformations of different spatial sizes. For instance, $\omega_{x0}=0.03$ will give deformation that will deform the whole brain hemisphere at a time, while $\omega_{x0}=0.3$ gives deformations that are 10 times smaller in spatial size and deforms regions of the size of individual lobes. The temporal frequency was chosen a little ad hoc, and future studies might benefit from finding plausible temporal frequencies from data. 

 After smoothing, the deformation flows were rescaled such that the standard deviation of the vector field $\sigma_v$ was $[0.1, 0.3, 1.0] \; \text{mm}/\text{step}$. This allows us to generate deformations of various magnitudes. For the smaller magnitudes it is likely that the deformations caused by the between-session changes of the subjects and imaging distortion are bigger than the synthetic deformation. However it is still interesting to generate the small deformations, since this gives an assessment of the practical limitation of the magnitude of deformations that can be recovered from time series of MRI imaging from different centers.  In total there were 12 different configurations of deformations, covering very small deformations to very large deformation.

\begin{figure}
    \centering
    \begin{subfigure}[b]{0.31\textwidth}
        \includegraphics[width=1.0\linewidth]{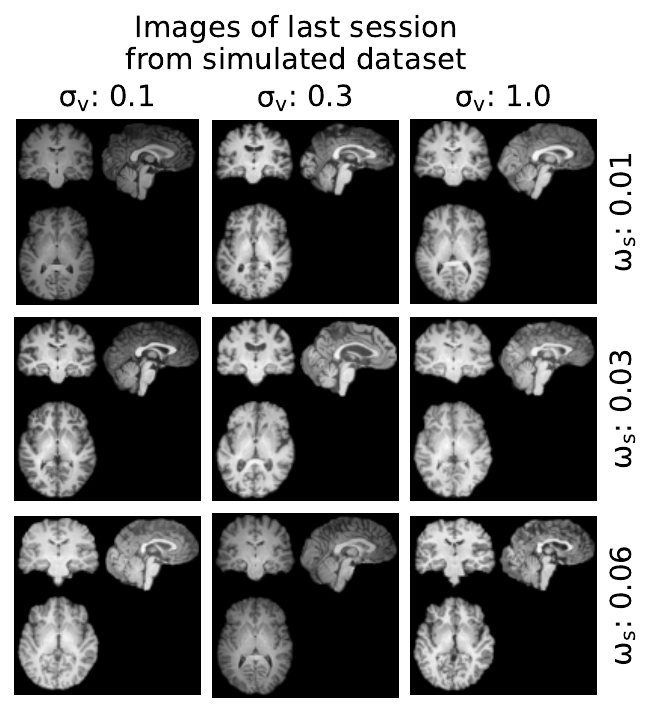}
    \end{subfigure}
    \begin{subfigure}[b]{0.31\textwidth}
        \includegraphics[width=1.0\linewidth]{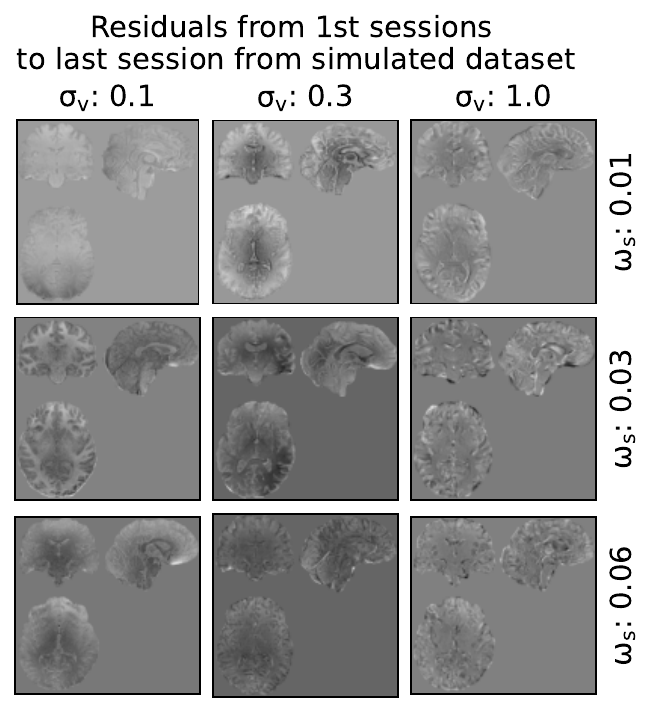}
    \end{subfigure}
    \begin{subfigure}[b]{0.31\textwidth}
        \includegraphics[width=1.0\linewidth]{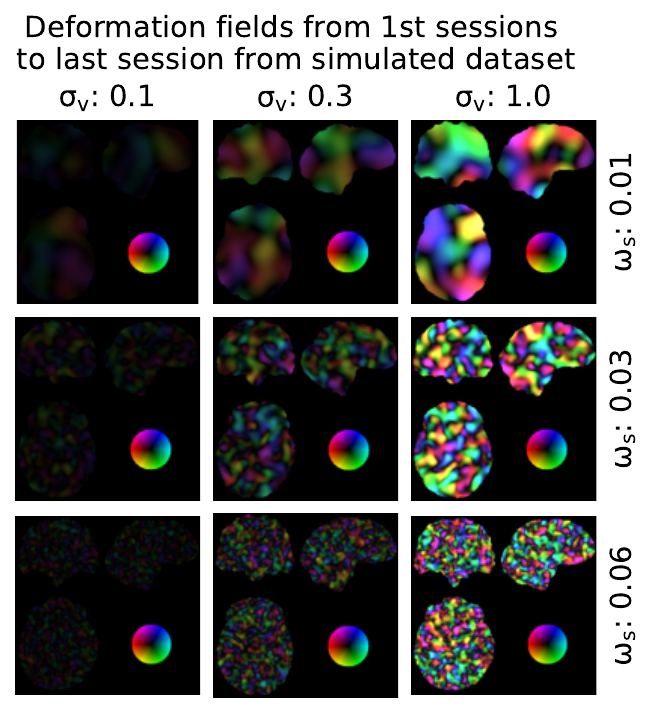}
    \end{subfigure}
    \caption{\textbf{Left:} The deformed images simulating the last session of the time series. The horizontal direction shows the different magnitudes of the deformation used while the vertical direction shows the different spatial smoothing scales of the deformations. \textbf{Middel:} The residual of the first and last session in the simulated longitudinal data. In the column to the right with the largest deformations one can see that not all edges are aligned due to the synthetic deformations. \textbf{Right:} Synthetic deformations fields applied to each of the images to the left. The color indicates the direction of the field orthogonal to the image plane and the brightness indicate the magnitude of deformations.}
    \label{fig:sim_data_vis}
\end{figure}

In Fig. \ref{fig:sim_data_vis} the synthetic deformations are illustrated with some examples from the dataset.

\subsubsection{Comparison to other registration methods}
\begin{figure}
    \centering
     \begin{subfigure}[b]{0.45\textwidth}
    \includegraphics[width=\linewidth]{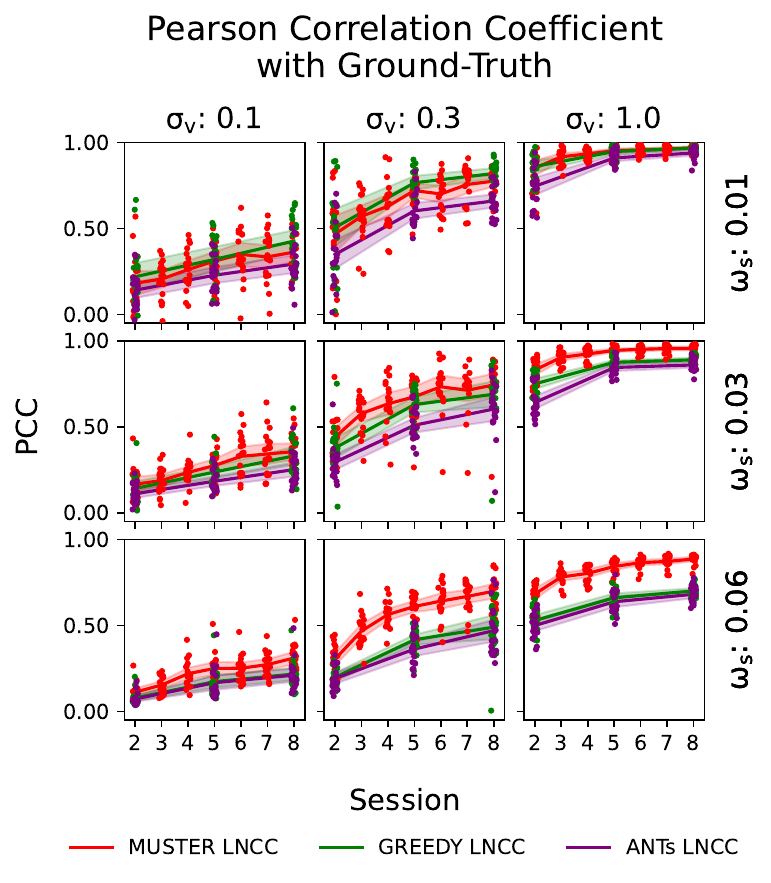}
    \end{subfigure}
    \begin{subfigure}[b]{0.45\textwidth}
    \includegraphics[width=\linewidth]{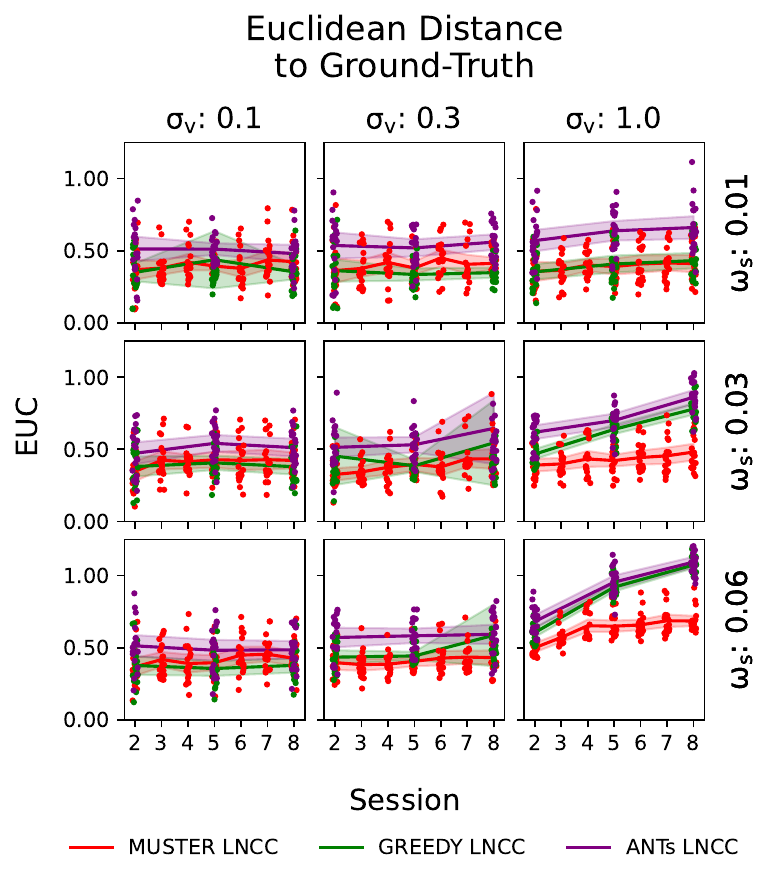}
    \end{subfigure}
    \begin{subfigure}[b]{0.45\textwidth}
    \includegraphics[width=\linewidth]{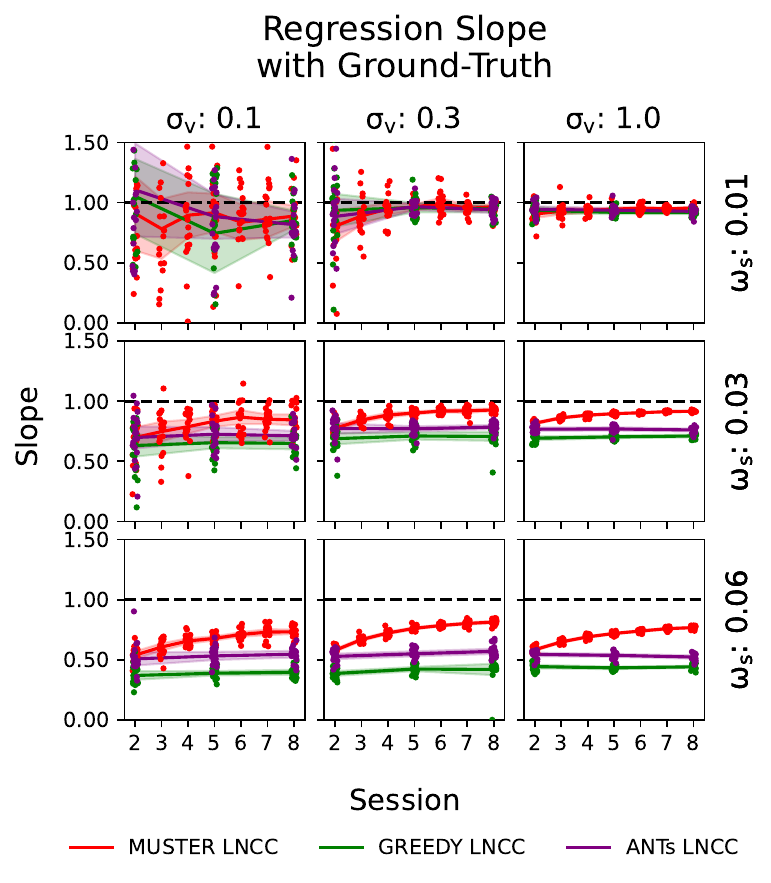}
    \end{subfigure}
    \caption{Performance metrics of of ANTs, GREEDY and MUSTER \textbf{Top Left:} The Pearson correlation coefficient between the true deformation fields and the estimated deformation fields. The x-axis shows the synthetic sessions, and y-axis the PCC between the estimated field and the true field for each session. The lines are the means for each session, and the shaded areas represent $95\%$ confidence interval for the mean. \textbf{Top Right:} The Euclidean distance between the estimated deformation field and the true deformation field. \textbf{Bottom:} The regression slope between the ground truth and the estimated deformation fields. }
    \label{fig:comparison_sim}
\end{figure}

\begin{table} 
\scriptsize
\begin{adjustbox}{center}
\setlength\tabcolsep{3.5pt}
\begin{tabular}{lc|ccc|ccc|ccc}
\toprule
 & metric & \multicolumn{3}{c}{EUC$\downarrow$} & \multicolumn{3}{c}{PCC$\uparrow$} & \multicolumn{3}{c}{Slope$\uparrow$} \\
 & method & ANTs & GREEDY & MUSTER & ANTs & GREEDY & MUSTER & ANTs & GREEDY & MUSTER \\
$\omega_s$ & $\sigma_v$ &  &  &  &  &  &  &  &  &  \\
\midrule
\multirow[t]{3}{*}{0.01} & 0.10 & $0.48${\tiny$\pm 0.03 $ } & $\mathbf{ 0.35 }${\tiny $\pm 0.02 $} & $0.42${\tiny$\pm 0.02 $ } & $0.29${\tiny$\pm 0.03 $ } & $\mathbf{ 0.43 }${\tiny $\pm 0.02 $} & $0.36${\tiny$\pm 0.01 $ } & $0.82${\tiny$\pm 0.06 $ } & $\mathbf{ 0.85 }${\tiny $\pm 0.05 $} & $\mathbf{ 0.89 }${\tiny $\pm 0.04 $} \\
 & 0.30 & $0.56${\tiny$\pm 0.04 $ } & $\mathbf{ 0.35 }${\tiny $\pm 0.01 $} & $0.41${\tiny$\pm 0.01 $ } & $0.67${\tiny$\pm 0.03 $ } & $\mathbf{ 0.82 }${\tiny $\pm 0.02 $} & $0.77${\tiny$\pm 0.01 $ } & $0.94${\tiny$\pm 0.02 $ } & $0.95${\tiny$\pm 0.02 $ } & $\mathbf{ 0.97 }${\tiny $\pm 0.01 $} \\
 & 1.00 & $0.66${\tiny$\pm 0.04 $ } & $\mathbf{ 0.43 }${\tiny $\pm 0.03 $} & $\mathbf{ 0.41 }${\tiny $\pm 0.02 $} & $0.94${\tiny$\pm 0.01 $ } & $\mathbf{ 0.97 }${\tiny $\pm 0.01 $} & $\mathbf{ 0.97 }${\tiny $\pm 0.00 $} & $0.93${\tiny$\pm 0.01 $ } & $0.92${\tiny$\pm 0.01 $ } & $\mathbf{ 0.96 }${\tiny $\pm 0.00 $} \\
\cline{1-11}
\multirow[t]{3}{*}{0.03} & 0.10 & $0.51${\tiny$\pm 0.02 $ } & $\mathbf{ 0.38 }${\tiny $\pm 0.02 $} & $0.42${\tiny$\pm 0.02 $ } & $0.25${\tiny$\pm 0.02 $ } & $0.33${\tiny$\pm 0.01 $ } & $\mathbf{ 0.36 }${\tiny $\pm 0.01 $} & $0.71${\tiny$\pm 0.02 $ } & $0.65${\tiny$\pm 0.02 $ } & $\mathbf{ 0.84 }${\tiny $\pm 0.03 $} \\
 & 0.30 & $0.66${\tiny$\pm 0.18 $ } & $\mathbf{ 0.57 }${\tiny $\pm 0.25 $} & $\mathbf{ 0.43 }${\tiny $\pm 0.03 $} & $0.60${\tiny$\pm 0.02 $ } & $\mathbf{ 0.69 }${\tiny $\pm 0.09 $} & $\mathbf{ 0.74 }${\tiny $\pm 0.02 $} & $0.78${\tiny$\pm 0.01 $ } & $0.71${\tiny$\pm 0.04 $ } & $\mathbf{ 0.93 }${\tiny $\pm 0.01 $} \\
 & 1.00 & $0.86${\tiny$\pm 0.03 $ } & $0.78${\tiny$\pm 0.01 $ } & $\mathbf{ 0.48 }${\tiny $\pm 0.03 $} & $0.86${\tiny$\pm 0.01 $ } & $0.89${\tiny$\pm 0.00 $ } & $\mathbf{ 0.96 }${\tiny $\pm 0.00 $} & $0.76${\tiny$\pm 0.01 $ } & $0.71${\tiny$\pm 0.01 $ } & $\mathbf{ 0.92 }${\tiny $\pm 0.01 $} \\
\cline{1-11}
\multirow[t]{3}{*}{0.06} & 0.10 & $0.49${\tiny$\pm 0.03 $ } & $\mathbf{ 0.38 }${\tiny $\pm 0.02 $} & $0.43${\tiny$\pm 0.01 $ } & $0.21${\tiny$\pm 0.01 $ } & $0.22${\tiny$\pm 0.01 $ } & $\mathbf{ 0.31 }${\tiny $\pm 0.01 $} & $0.55${\tiny$\pm 0.02 $ } & $0.40${\tiny$\pm 0.01 $ } & $\mathbf{ 0.74 }${\tiny $\pm 0.01 $} \\
 & 0.30 & $0.59${\tiny$\pm 0.03 $ } & $0.59${\tiny$\pm 0.18 $ } & $\mathbf{ 0.43 }${\tiny $\pm 0.02 $} & $0.46${\tiny$\pm 0.04 $ } & $0.49${\tiny$\pm 0.03 $ } & $\mathbf{ 0.70 }${\tiny $\pm 0.01 $} & $0.57${\tiny$\pm 0.02 $ } & $0.42${\tiny$\pm 0.05 $ } & $\mathbf{ 0.81 }${\tiny $\pm 0.01 $} \\
 & 1.00 & $1.09${\tiny$\pm 0.02 $ } & $1.07${\tiny$\pm 0.01 $ } & $\mathbf{ 0.69 }${\tiny $\pm 0.02 $} & $0.68${\tiny$\pm 0.02 $ } & $0.70${\tiny$\pm 0.00 $ } & $\mathbf{ 0.89 }${\tiny $\pm 0.01 $} & $0.52${\tiny$\pm 0.01 $ } & $0.44${\tiny$\pm 0.01 $ } & $\mathbf{ 0.77 }${\tiny $\pm 0.01 $} \\
\cline{1-11}
\bottomrule
\end{tabular}

\end{adjustbox}
\caption{Performance metrics of the last session in the simulated dataset.  \textbf{Bold} numbers mark that there is a probability above $5\%$ that this model is the best according to the metric. See Appendix \ref{se:model_comp} for details. $\pm$ indicates two standard deviations of the performance estimate.}
\label{tab:Long_reg}
\end{table}
We compared MUSTER to two other popular registration methods: ANTs SyN \cite{avants2009advanced} and Greedy \cite{joshi2004unbiased, yushkevich2019}. ANTs was chosen since it is a popular software for image registration. Greedy has been shown to be state-of-the-art for deformable registration \cite{jena2024deep}. For both algorithms we used LNCC with a window size of $3\times 3 \times 3$. For ANTs SyN we used the default setting found in ANTspyx \cite{tustison2021antsx}, and we matched the same parameters for Greedy. 
In the first synthetic experiment we used MUSTER to find the deformations between all 8 session. Greedy and ANTs SyN performed pairwise registration between session 1 and session $[2, 4, 7]$. Only a subset of the sessions was used to save compute time as ANTs is quite slow to run. The performance metrics can be seen in Tab. \ref{tab:Long_reg} and in Fig. \ref{fig:comparison_sim}.
All metrics were calculated within the brain volume only. 

We see that for deformations that have small spatial extent, MUSTER outperforms both ANTs SyN and Greedy, however, Greedy does have slightly better performance for spatially large deformations.

\subsubsection{Number of images in time series}
We ran the longitudinal registration on the synthetic data using MUSTER with a different number of imaging sessions. We investigated how adding intermediate images in a series changes the estimation of the deformation from the first image to the last image.

\begin{figure}
    \centering
     \begin{subfigure}[b]{0.45\textwidth}
    \includegraphics[width=\linewidth]{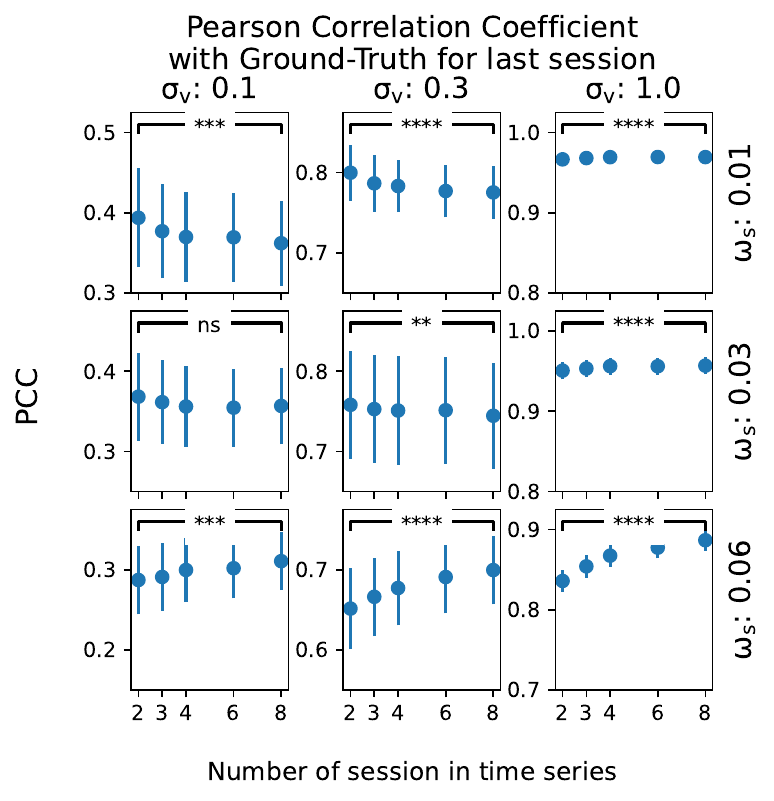}
    \end{subfigure}
    \begin{subfigure}[b]{0.45\textwidth}
    \includegraphics[width=\linewidth]{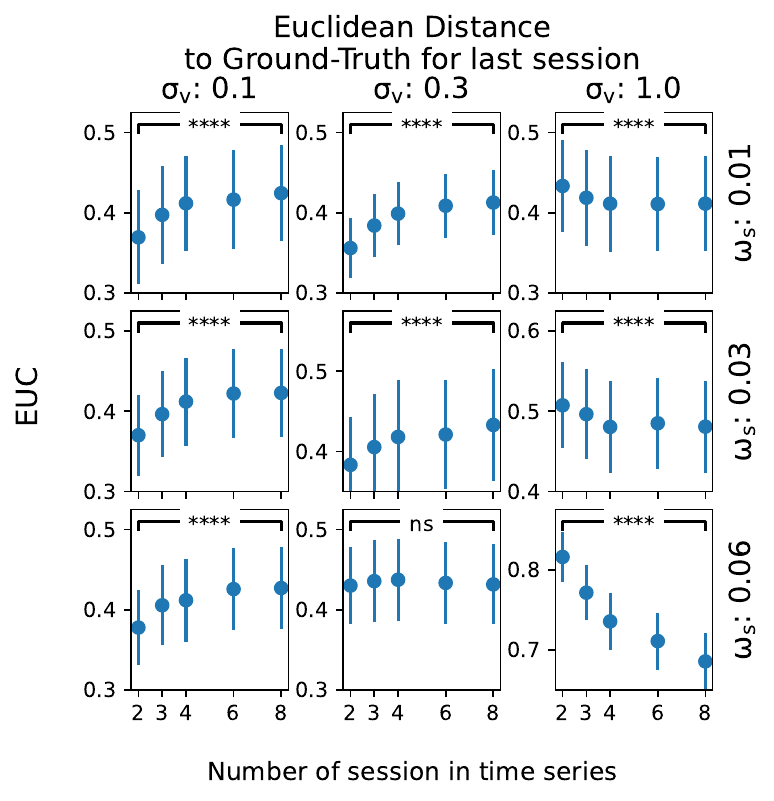}
    \end{subfigure}
    \begin{subfigure}[b]{0.45\textwidth}
    \includegraphics[width=\linewidth]{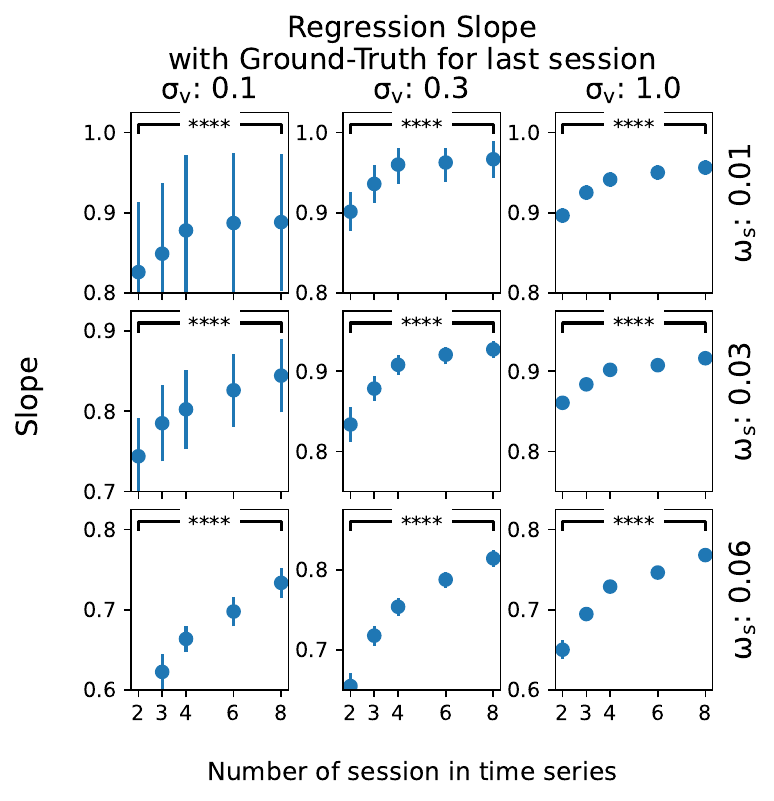}
    \end{subfigure}
    \caption{Investigating how the number of images in a time series impacts the recovery of the deformations from first time point to the last time point. The x-axis represent the number of simulated images in the time series and the y-axis is the mean performance metric between the groudn truth and the deformation beween image session 0 and 8. The 95\% confidence interval of the mean is indicated with errorbars. A paired t-test is done between using 2 sessions and 8 sessions. p value is indicated with: $p\leq*:0.05, \; **:0.01, \;  **:0.001,\; ***:0.0001$}
    \label{fig:comparison_num_session}
\end{figure}

In Fig. \ref{fig:comparison_num_session} the performance metrics for the last sessions are plotted with respect to the total number of sessions in the time series.
We see that for spatially small and large deformations, PCC and Euclidean distance indicate that including more images is beneficial to increased performance. For the other configurations, we see small or even a small negative benefit to include more images. This might have to do with how MUSTER is regularized. With fewer intermediate deformations between first and last image, there is more regularization on the flow fields, resulting in smoother deformations, which can be of benefit in deformation with spatially large extent. From the slope estimates we see that adding more images always decrease the bias of the deformation field. 



\subsection{Sensitivity to clinical outcomes}
In this section MUSTER is applied to a clinical application on a real-life dataset. Inspired by \citet{ashburner2013symmetric} we used MUSTER to evaluate the expansion and contractions of brain regions of healthy controls and Alzheimer's disease (AD) patients in the ADNI dataset and correlate this with the change in cognitive function.

\subsubsection{Experiment setup}
We randomly selected 249 subjects with multiple time points from the ADNI database. \url{adni.loni.usc.edu}. The ADNI was launched in 2003 as a public-private partnership, led by Principal Investigator Michael W. Weiner, MD. The primary goal of ADNI has been to test whether serial MRI, positron emission tomography (PET), other biological markers, and clinical and neuropsychological assessment can be combined to measure the progression of mild cognitive impairment (MCI) and early Alzheimer’s disease.

\begin{figure}
    \centering
    \includegraphics[width=1.0\linewidth]{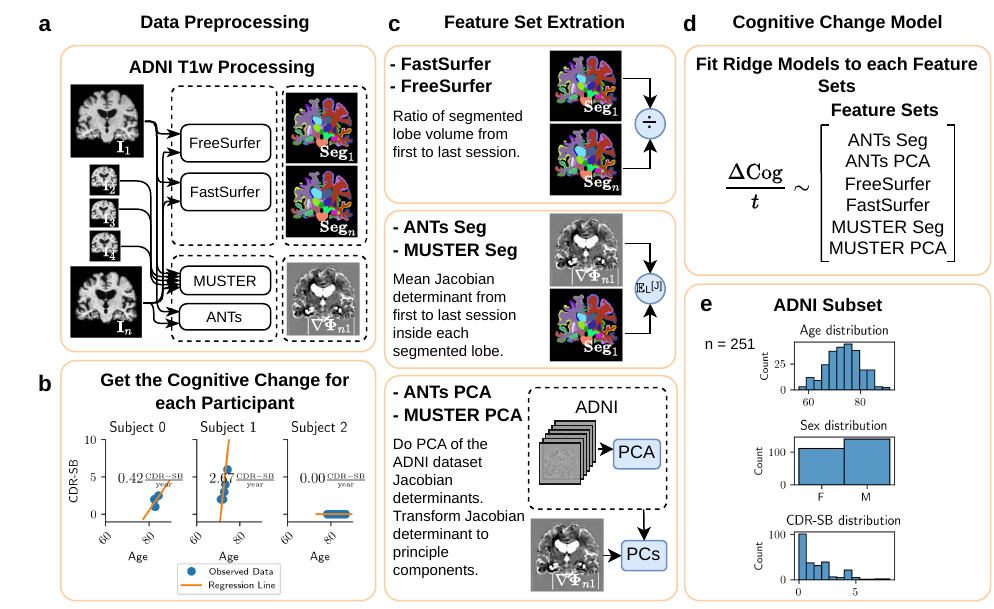}
    \caption{Overview of the comparison of methods for processing longitudinal brain MRI using ANTs, MUSTER, FreeSurfer and FastSurfer. \textbf{a: } The first and last scan of each participant was segmented using FreeSurfer longitudinal and FastSurfer. ANTs and MUSTER were used to estimate the Jacobian determinant from the first session to the last session. MUSTER also uses the intermediate time points. \textbf{b:} The slope of MMSE and CDR-SB were estimated using linear regression. The plot shows the CDR-SB over time for three selected participants. \textbf{c:} Features are extracted in three ways: FastSurfer and FreeSurfer estimates of expansion ratio is created for each lobe by division. Second, the mean ANTs and MUSTER Jacobian determinants are calculated for each of the segmented areas as determined by FastSurfer. Third, a PCA with 32 components of all Jacobian determinants in the ADNI sample is used to extract the features directly from the Jacobian determinants. \textbf{d:} Ridge regression models are fitted to the change in cognitive score with the features in c as inputs. \textbf{e:} Age, sex and CDR-SB distribution of the ADNI subset used. }
    \label{fig:ADNIexp}
\end{figure}

See Fig. \ref{fig:ADNIexp} for an graphical overview of the experimental setup. We used MUSTER to find the deformation between the first session and all following sessions of each subject. We then calculated the Jacobian determinant of the deformations to determine the average monthly expansion and contraction of every voxel. We extracted features from the Jacobian determinant maps using two different methods. The first method used FastSurfer \citep{henschel2020fastsurfer} to segment the T1-weighted images, and for each region the mean Jacobian determinant was calculated. 
In the second method the Jacobian determinant maps were transformed to MNI space \cite{fonov2009unbiased, fonov2011unbiased} using ANTs. We then used principal component analysis (PCA) to extract the 32 principal components (PCs) of the Jacobian determinant of the ADNI dataset. The PCs of each subject was used as features. 
For both of these two methods we followed the same procedure using ANTs SYN for the Jacobian determinants as a comparison, except that only the first and last session was used since ANTs uses a parwise registration method. 

We also compare to segmentation based feature extraction. FreeSurfer longitudinal \cite{reuter2012within} and FastSurfer \cite{henschel2020fastsurfer} was used to segment the first and last session. The volumetric ratio of the last to first session was calculated for each region. The volumetric ratios of a region should be equivalent to the mean Jacobian determinants in that region given that the segmentation and the registrations are accurate. 

This results in 6 sets of features: \textit{ANTs Seg} and \textit{MUSTER Seg}: mean Jacobian determinant in regions, \textit{ANTs PCA} and \textit{MUSTER PCA}: PC of Jacobian determinant, \textit{FreeSurfer} and \textit{FastSurfer}: Ratio of the segmented volume of each region.

For each subject, we calculated the slope of the Clinical Dementia Rating sum of boxes (CDR-SB) and for Mini-Mental Scale Examination (MMSE) for all sessions using linear regression. We use this slope as a measure of cognitive change. See Fig. \ref{fig:ADNIexp}b for some examples. The CDR-SB score is a popular metric for pharmaceutical trials applied to Alzheimer's disease and a metric of a subjects deviation from normal cognitive function \cite{budd2022two}. MMSE serves  a similar function and is also wieldy used cognitive metric for people with dementia \cite{arevalo2021mini}.

Ridge regression models were fitted with a hyperparameter search for each feature set relating them to the cognitive change variables. We measured the performance of each model using Root Mean Squared Error (RMSE), Mean Absolute Error (MAE), and PCC. For all procedures nested 10 fold cross validation was used, and all reported metrics are from the test folds. We calculated the probability of each model being the best for each metric using a Bayesian mixed model. See Appendix \ref{se:model_comp} for an overview of this method.

\subsubsection{Results}
The performance of the models is displayed in Table \ref{tab:adni_long_reg}. The MUSTER PCA performed overall best on all metrics, with a PCC of 0.56 for the CDR-SB model and 0.64 for the MMSE model. Compare this to for instance Freesurfer longitudinal with a PCC of 0.43 for CDR-SB and 0.48 for MMSE. MUSTER PCA was always in the category of methods with a chance of being the best method for all metrics, however as seen in Table \ref{tab:adni_long_reg} there where multiple models with a probability $p>0.05$ of being the best model.  

\begin{table} 
\scriptsize
\begin{adjustbox}{center}
\setlength\tabcolsep{3.5pt}
\begin{tabular}{l|ccc|ccc}
\toprule
 & \multicolumn{3}{c}{CDR-SB} & \multicolumn{3}{c}{MMSE} \\
 & RMSE$\downarrow$ & MAE$\downarrow$ & PCC$\uparrow$ & RMSE$\downarrow$ & MAE$\downarrow$ & PCC$\uparrow$ \\
\midrule
ANTs Seg & $0.99${\tiny$\pm 0.02 $ } & $0.65${\tiny$\pm 0.03 $ } & $0.42${\tiny$\pm 0.09 $ } & $1.86${\tiny$\pm 0.07 $ } & $1.18${\tiny$\pm 0.04 $ } & $0.53${\tiny$\pm 0.06 $ } \\
ANTs PCA & $\mathbf{ 0.92 }${ \tiny $\pm 0.05 $} & $\mathbf{ 0.58 }${ \tiny $\pm 0.03 $} & $\mathbf{ 0.49 }${ \tiny $\pm 0.06 $} & $1.75${\tiny$\pm 0.05 $ } & $1.04${\tiny$\pm 0.06 $ } & $0.58${\tiny$\pm 0.02 $ } \\
FreeSurfer & $0.99${\tiny$\pm 0.09 $ } & $0.65${\tiny$\pm 0.05 $ } & $0.43${\tiny$\pm 0.10 $ } & $1.88${\tiny$\pm 0.16 $ } & $1.15${\tiny$\pm 0.09 $ } & $0.48${\tiny$\pm 0.13 $ } \\
FastSurfer & $\mathbf{ 0.95 }${ \tiny $\pm 0.11 $} & $0.64${\tiny$\pm 0.06 $ } & $\mathbf{ 0.41 }${ \tiny $\pm 0.19 $} & $\mathbf{ 2.24 }${ \tiny $\pm 0.77 $} & $1.33${\tiny$\pm 0.31 $ } & $0.40${\tiny$\pm 0.18 $ } \\
MUSTER Seg & $0.94${\tiny$\pm 0.05 $ } & $0.60${\tiny$\pm 0.04 $ } & $0.44${\tiny$\pm 0.09 $ } & $\mathbf{ 1.74 }${ \tiny $\pm 0.12 $} & $1.09${\tiny$\pm 0.07 $ } & $\mathbf{ 0.57 }${ \tiny $\pm 0.10 $} \\
MUSTER PCA & $\mathbf{ 0.89 }${ \tiny $\pm 0.03 $} & $\mathbf{ 0.56 }${ \tiny $\pm 0.04 $} & $\mathbf{ 0.56 }${ \tiny $\pm 0.07 $} & $\mathbf{ 1.67 }${ \tiny $\pm 0.06 $} & $\mathbf{ 0.97 }${ \tiny $\pm 0.03 $} & $\mathbf{ 0.64 }${ \tiny $\pm 0.06 $} \\
\bottomrule
\end{tabular}

\end{adjustbox}
\caption{Comparison of methods for explaining change in MMSE and CDR-SB from volumetric data. \textbf{Bold} numbers mark that there is a probability above $5\%$ that this model is the best according to the metric. $\pm$ indicates two standard deviation of the performance estimate.}
\label{tab:adni_long_reg}
\end{table}

\section{Discussion}
The results demonstrate that MUSTER effectively estimates deformations in longitudinal imaging time series, producing clinically relevant volumetric changes. These changes exhibit explanatory power comparable to widely used segmentation tools when relating volumetric measurements to cognitive performance.

A significant advantage of MUSTER is its flexibility in handling various image similarity metrics. Unlike Geodesic Regression, which relies on a Riemannian metric and is therefore incompatible with metrics such as SiLNCC and Mutual Information, MUSTER is agnostic to the choice of similarity metric, broadening its applicability.

Another strength of MUSTER is its computational efficiency. By utilizing GPU acceleration and and extending the Log-Euclidean framework to longitudinal series, the algorithm accommodates large datasets with limited computational resources. This efficiency makes MUSTER a practical choice for both research and clinical applications.

Despite these promising results, several limitations warrant further discussion. One key limitation is the reliance on hyperparameter tuning, particularly for the regularization terms. As with other registration methods (e.g., ANTs SyN, Greedy), the choice of hyperparameters significantly impacts performance. While an exhaustive hyperparameter search is ideal, its computational demands made this impractical. Therefore, we relied on default settings, which may not represent optimal configurations.

In our experiments, a combination of parametric regularization (via smoothing) and a regularizing loss function was used to improve tuning. Smoothing ensured a baseline level of smoothness in the flow field, while the loss function maintained smoothness in regions with limited information. While this approach facilitated tuning, it also highlights the need for further investigation into alternative regularization techniques. 

Another limitation of MUSTER is its assumption that anatomical changes can be fully described by diffeomorphic deformations. While standard in medical image registration, this assumption may not capture scenarios involving tissue appearance or disappearance, such as tumor growth or surgical resection. Future work could extend MUSTER to accommodate such cases by incorporating models that allow for changes to intensities over time. SiLNCC is only somewhat robust to changes in intensities, and will produce biased results when the when the changes in a region is not well described by a linear model which is assumed for SiLNCC.

\section{Conclusion}
In this paper, we introduced MUSTER, a novel algorithm for longitudinal registration of medical images that effectively incorporates multiple imaging sessions to enhance registration precision. By composing consecutive deformations and leveraging both rigid and non-linear registration, MUSTER adds temporal constraints that guide the estimated deformation fields along plausible anatomical trajectories. This approach addresses limitations of conventional pairwise registration by utilizing the additional information inherent in multiple time points, leading to improved robustness. By using SiLNCC as an alternative to the cross-correlation as similarity metric, our method is robust against imaging artifacts such as noise, contrast changes, and bias fields.

Our experimental results demonstrate the effectiveness of MUSTER in both synthetic and real-world datasets. In synthetic tests, MUSTER outperformed established registration methods like ANTs SyN and Greedy, in scenarios involving small spatial deformations, by more accurately recovering ground truth deformations. When applied to the ADNI dataset, MUSTER successfully identified patterns of neurodegeneration from T1-weighted MRI scans and these patterns ccorrelates with changes in cognitive function as measured by CDR-SB and MMSE scores. These findings highlight its potential for clinical applications and longitudinal studies of neurodegenerative diseases.

In summary, MUSTER offers a robust and computationally efficient framework for analyzing anatomical changes over time. Its ability to leverage multiple imaging sessions for improved registration precision makes it a valuable tool for longitudinal studies and clinical workflows, where detecting and characterizing subtle tissue changes is critical.
\clearpage

\section*{Data and Code Availability}
The code can be accessed at \url{https://github.com/CRAI-OUS/MUSTER}. The ADNI dataset can be accessed by request from \url{https://adni.loni.usc.edu/data-samples/adni-data/#AccessData}. The LCBC Traveling Brains can be accessed upon request to LCBC, and will be released as a dataset on a future timepoint. 

\section*{Author Contributions}

EG conceived the method, developed the code, performed the experiments, and drafted the initial manuscript. DS enhanced mathematical rigor and conducted additional experiments. ØS supported statistical analysis. IA preprocessed the data. EHL, BM, AT, TS, PG, AB and AF provided expertise in medical imaging, assessed clinical utility, and contributed to validating the method. AF also provided funding and contributed in a leadership and advisory role. All named authors reviewed the manuscript. 

\section*{Funding}
The project was supported by a grant from the South-Eastern Norway Regional Health Authority (HSØ- 2021079).

\section*{Ethical Statement}
All participants for the LCBC Traveling Brains dataset have informed constent to the study, and all data has been anonymized. The study has been reviewed by an ethics committee from the Regional Committees for Medical Research Ethics South East Norway. 

The clinical experiments of ADNI have been approved by the ethics board selected by the participating institutes of ADNI. The Office for Human Research Protections (OHRP) has reviewed and approved each ethics board. Informed consent from all participants in ADNI has been conducted in accordance with US 21 CFR 50.25, the Tri-Council Policy Statement: Ethical Conduct of Research Involving Humans and the Health Canada and ICH Good Clinical Practice. All methods in the current study were carried out following the guidelines and regulations of ADNI.

\section*{Declaration of Competing Interests}

EHL is the CSO and a major shareholder in baba.vision. The remaining authors declare no competing interests.

\section*{Acknowledgements}
Data collection and sharing for this project was funded by the Alzheimer's Disease Neuroimaging Initiative (ADNI) (National Institutes of Health Grant U01 AG024904) and DOD ADNI (Department of Defense award number W81XWH-12-2-0012). ADNI is funded by the National Institute on Aging, the National Institute of Biomedical Imaging and Bioengineering, and through generous contributions from the following: AbbVie, Alzheimer’s Association; Alzheimer’s Drug Discovery Foundation; Araclon Biotech; BioClinica, Inc.; Biogen; Bristol-Myers Squibb Company; CereSpir, Inc.; Cogstate; Eisai Inc.; Elan Pharmaceuticals, Inc.; Eli Lilly and Company; EuroImmun; F. Hoffmann-La Roche Ltd and its affiliated company Genentech, Inc.; Fujirebio; GE Healthcare; IXICO Ltd.; Janssen Alzheimer Immunotherapy Research \& Development, LLC.; Johnson \& Johnson Pharmaceutical Research \& Development LLC.; Lumosity; Lundbeck; Merck \& Co., Inc.; Meso Scale Diagnostics, LLC.; NeuroRx Research; Neurotrack Technologies; Novartis Pharmaceuticals Corporation; Pfizer Inc.; Piramal Imaging; Servier; Takeda Pharmaceutical Company; and Transition Therapeutics. The Canadian Institutes of Health Research is providing funds to support ADNI clinical sites in Canada. Private sector contributions are facilitated by the Foundation for the National Institutes of Health \url{www.fnih.org}. The grantee organization is the Northern California Institute for Research and Education, and the study is coordinated by the Alzheimer’s Therapeutic Research Institute at the University of Southern California. ADNI data are disseminated by the Laboratory for Neuro Imaging at the University of Southern California.

\printbibliography 

\newpage
\setcounter{section}{0}
\renewcommand{\thesection}{\Alph{section}}
\addcontentsline{toc}{section}{Appendix}
\section{Appendix}
\subsection{Gaussian Smoothing}
\label{se:gaussian_smoothing}
We used a Gaussian filter multiple times in this paper. This was implemented in the Fourier domain using fast Fourier transformations.

The method for creating the deformations can be summarized as: \begin{subequations}
    \begin{equation}
        \bm n^*(\bm \omega) = \mathscr{F}\{\bm n(\bm x) \} (\bm \omega),
    \end{equation}
    \begin{equation}
        \bm v^*(\bm \omega) =  e^{-\frac{1}{2} \left( \frac{\bm \omega}{\bm{\omega}_{s}} \right)^2 }  \bm n^*(\omega_t, \bm \omega_s),
    \end{equation}
     \begin{equation}
        \bm v(\bm x) = \mathscr{F}^{-1}\{ \bm v^*(\bm \omega_s) \} (\bm x),
    \end{equation}
 \end{subequations}
where:
\begin{itemize}
    \item $\bm n(\bm x)$ is the input field
    \item $\bm x$ is the spatial position
    \item $ \bm \omega$ is the spatial complex frequency in the Fourier domain
    \item $\bm \omega_s $ is the spatial frequency filtering constant 
    \item $\mathscr{F}$ and $\mathscr{F}^-1$ is the Fourier transform and it's inverse.
    \item $ \bm v(\bm x)$ is the smooth output field 
\end{itemize}

\subsection{Generation of Synthetic dataset}
Here we describe in details how the synthetic deformations was generated. 

First a random field was created by drawing from the a Gaussian distribution at the resolution of the images. We choose 12 intermediate integration steps between each session, giving $T=12\cdot 8$.    

\begin{equation}
    \bm n(t, \bm x) \sim N(\bm 0, \bm 1) \in \mathbb{R}^{T \times 3 \times W \times H \times D}
\end{equation}
The noise field was then smoothed using the method described in Appendix \ref{se:gaussian_smoothing}. The time dimension and the spatial dimension was smoothed with different smoothing constants as described in Section \ref{se:syn_data} giving a smooth but unscaled flow field $\bm v(t, \bm x)$. This field was scaled such that the standard divination of the flow field matched $\omega_v$
 \begin{equation}
    \bm v_s(t, \bm x) =  v(t, \bm x) \frac{\sigma_v}{\sqrt{\text{Var}(v(t, \bm x))}}
\end{equation}
The deformations fields are obtained by integrating the flow field:
\begin{equation}
    \bm \Phi(t, \bm x) = \int_0^t \bm v_s(\tau, \bm x) \circ \bm \Phi(\tau, \bm x) d \tau + \bm \Phi_{\mathbb{I}}
\end{equation}
We used the method from \cite{staniforth1991semi} to integrate the flow field. 

\subsection{Expectation of LNCC}
\label{apx:Expectation_lncc}
In this section we show how the the approximation of the expectation of the LNCC is calculated. Some simplifying assumptions are made and we show in a simulation that these simplification can be neglected in realistic scenarios. 

For convenience we repeat the definition of LNCC:
\begin{equation}
    \label{eq:lncc_appendix}
    \text{LNCC}_R = 1 - \frac{1}{|R|} \frac{\sum_{\bm{x} \in R} \bar{\bm I}_{i\bm{x}} \bar{\bm I}_{j\bm{x}}}{\sqrt{S_{i}^2 S_{j}^2}}. \tag{\ref{eq:LNCC} repeated}
\end{equation}

Where $\bar{\bm I}_{i\bm{x}} = \bm I_{i\bm{x}} - 1/|R|\sum_{\bm{x}' \in R}\bm I_{i\bm{x}'}$ and  $S_{i}^2 = 1/|R|\sum_{\bm{x}' \in R} (\bar{\bm I}_{i\bm{x}'R})^2$ 

We assume as before that within a small enough region there is a linear relationship between the intensities of the two images and that the errors $\epsilon_{\bm x}$ are Gaussian. 

\begin{equation}
\label{eq:lin_relation_appendix}
\bm I_{j\bm{x}} = a_{R} \bm I_{i\bm{x}} + b_{R} + \bm \epsilon_{\bm x}   \tag{\ref{eq:lin_relation} repeated}
\end{equation} 

Inserting Eq. \ref{eq:lin_relation} into Eq. \ref{eq:LNCC} we obtain:

\begin{equation}
    E_\epsilon[\text{LNCC}_R] = E_\epsilon\left[1 - \frac{1}{\sqrt{\sum_{\bm x \in R} \bar{\bm I}_{\bm x i}^2 }} \frac{\sum_{\bm x \in R} \bar{\bm I}_{\bm x i}(\epsilon_{\bm x} - \bar \epsilon) +  a_{R}\sum_{\bm x \in R} \bar{\bm I}_{\bm x i}^2 }{\sqrt{2a_{R} \sum_{\bm x \in R}  \bar{\bm I}_{\bm x i}(\epsilon_{\bm x} - \bar \epsilon) + \sum_{\bm x \in R} a_{R}^2 \bar{\bm I}_{\bm x i}^2 +  (\epsilon_{\bm x} - \bar \epsilon)^2}} \right] 
\end{equation}

For a sufficiently large region $R$,  $\sum_{\bm x \in R} \bar{\bm I}_{\bm x i}(\epsilon_{\bm x} - \bar \epsilon) \approx 0$. We can thus write

\begin{equation}
    \label{eq:lncc_first_apporx}
    E_\epsilon[\text{LNCC}_R] \approx E_\epsilon\left[1 - \frac{1}{\sqrt{\sum_{\bm x \in R} \bar{\bm I}_{\bm x i}^2 }} \frac{a_{R}\sum_{\bm x \in R}  \bar{\bm I}_{\bm x i}^2 }{\sqrt{\sum_{\bm x \in R} a_{R}^2 \bar{\bm I}_{\bm x i}^2 + \sum_{\bm x \in R} (\epsilon_{\bm x} - \bar \epsilon)^2}} \right]\\
    = 1 - E_{e^2} \left[ \frac{C}{\sqrt{C^2+e^2}}\right],
\end{equation}
where $C = a_{R} \frac{1}{|R|}\sum_{\bm x \in R} \bar{\bm I}_{\bm x i}²$ and $e^2 =  \frac{1}{|R|}\sum_{\bm x \in R} (\epsilon_{\bm x} - \bar \epsilon)^2$. 

$|R|e^2/\sigma^2$ is Chi-squared distributed with $|R|-1$ degrees of freedom. Again under the assumptions of large enough sample, a Chi-squared distribution can be approximated by a Gaussian distribution. This gives the following:

\begin{equation}
     E[\text{LNCC}_R] \approx 1- E_{|R|e^2\sim\sigma^2\chi^2_{|R|}}\left[ \ \frac{C}{\sqrt{C^2 +  e^2}}\right] \approx 1- E_{u \sim N(0, 1)} \left[\frac{C}{\sqrt{C^2 +  u \sigma^2\sqrt{\frac{2}{|R|}} + \sigma^2}} \right]
\end{equation}
Finally when $R$ is sufficiently large we can remove the stochastic variable $u$ and we arrive at Eq. \ref{eq:E_lncc}:

\begin{equation}
     E[\text{LNCC}_R]  \approx 1-  E_{u \sim N} \left[\frac{C}{\sqrt{C^2 + \sigma^2}} \right] = 1- \frac{1}{\sqrt{1 + \frac{\sigma^2}{a_{R}S_i^2}}}
     \tag{\ref{eq:E_lncc} repeated}
\end{equation}

\begin{figure}
    \centering
    \includegraphics[width=1\linewidth]{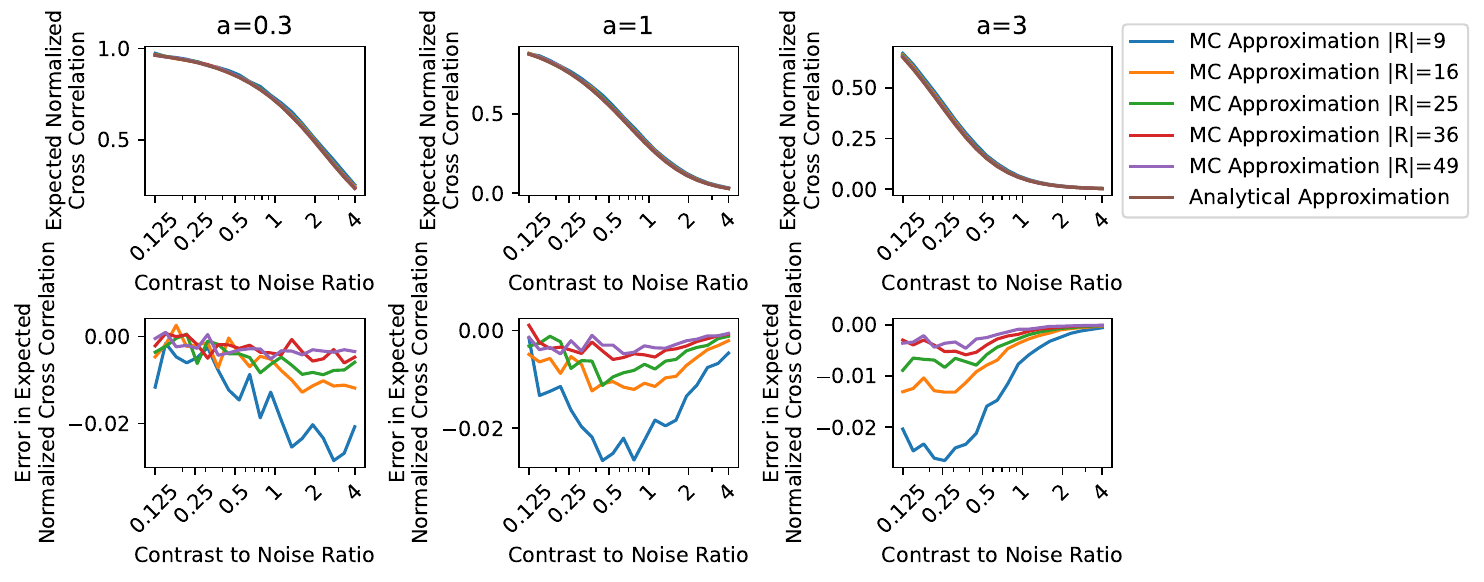}
    \caption{Estimates of $E[\text{LNCC}]$ as a function of CNR for three values of $a_{R}$. In the top plots Monte Carlo estimates are shown for different values of $|R|$ along with Eq. \ref{eq:E_lncc}. Note that the MC estimates are so close to Eq. \ref{eq:E_lncc} that it is hard to separate the lines. The residuals between the MC estimates and the Eq. \ref{eq:E_lncc} is plotted below.}
    \label{fig:expected_ncc}
\end{figure}

We evaluate the analytical approximation Eq. \ref{eq:E_lncc} of the $E[\text{LNCC}_R] $ by comparing to an Monte Carlo estimate of the expected value. The result can be seen in Fig. \ref{fig:expected_ncc}. We see that even for the smallest practical 2D kernal with $3\times3=9$ voxels, Eq. \ref{eq:E_lncc} is still a god approximation with only a maximum bias $\sim 0.02$. For 3D images the smallest practical kernal is $3\times3\times3=27$ and there the error is even smaller. We therefore conclude that for our analysis of LNCC in Section \ref{se:loss_func} Eq. \ref{eq:E_lncc} is a sufficient approximation. 

\subsection{Model Comparison}
\label{se:model_comp}
When comparing models, we used $N$ folds(or datasamples) and $M$ different models. We use a Baysian framework for the statistical analysis. We assume that the folds might be of different difficulty, reflected in different mean performances. We also assume that the mean and variance of the performance may vary between models. Building on this, we assume that the scores are not subject to ceiling or floor effects—that is, they are sufficiently far from their minimum and maximum possible of each score function—allowing us to model the error terms as Gaussian. This gives the following model:

\begin{equation}
    \begin{aligned}
        s_{ij} &= \beta_{m_i} + \beta_{f_j} + \epsilon_{ij}, \\
        \epsilon_{ij} &\sim N(0, \sigma_{m_i}^2),
    \end{aligned}
\end{equation}
where:
\begin{itemize}
    \item \( s_{ij} \) is the score for model \( m_i \) in fold \( f_j \),
    \item \( \beta_{m_i} \) is the effect of model \( m_i \),
    \item \( \beta_{f_j} \) is the effect of fold \( f_j \),
    \item \( \epsilon_{ij} \) is the error term with variance depending on the model.
\end{itemize}

We used a weak priors on $\beta_{m_i}$ and $\beta_{f_j}$. We constrain $\sum_{j=0}^N \beta_{f_j}=0$ such that $\beta_{m_i}$ captures the offset from zero.

\begin{subequations}
    \begin{equation}
        \beta_{m_i} \sim N(0,1)
    \end{equation}
    \begin{equation}
        \beta_{f_j}' \sim N(0, 0.3)
    \end{equation}
    \begin{equation}
        \beta_{f_j} = \beta_{f_j} - \frac{1}{N}\sum_{j'=0}^N \beta_{f_j'}
    \end{equation}
\end{subequations}

We fit this model to each score type using Markov Chain Monte Carlo using PyMC\cite{abril2023pymc}, obtaining estimates for the posterior distributions of \( \beta_{m_i} \). We then draw samples from the posterior distributions of the model effects and, report the mean and 2 standard deviations of the posterior for each model. The probability of model \( m_i \) being the best is estimated as the proportion of samples where \( \beta_{m_i} \) is greater than all other \( \beta_{m_j} \) for \( j \neq i \). In the result tables, we highlight all models with \( p(m_i \text{ is best}) > 0.05 \). 

\subsection{Additional Plots of Model Performance on ADNI}

\begin{figure}
    \centering
    \includegraphics[width=\linewidth]{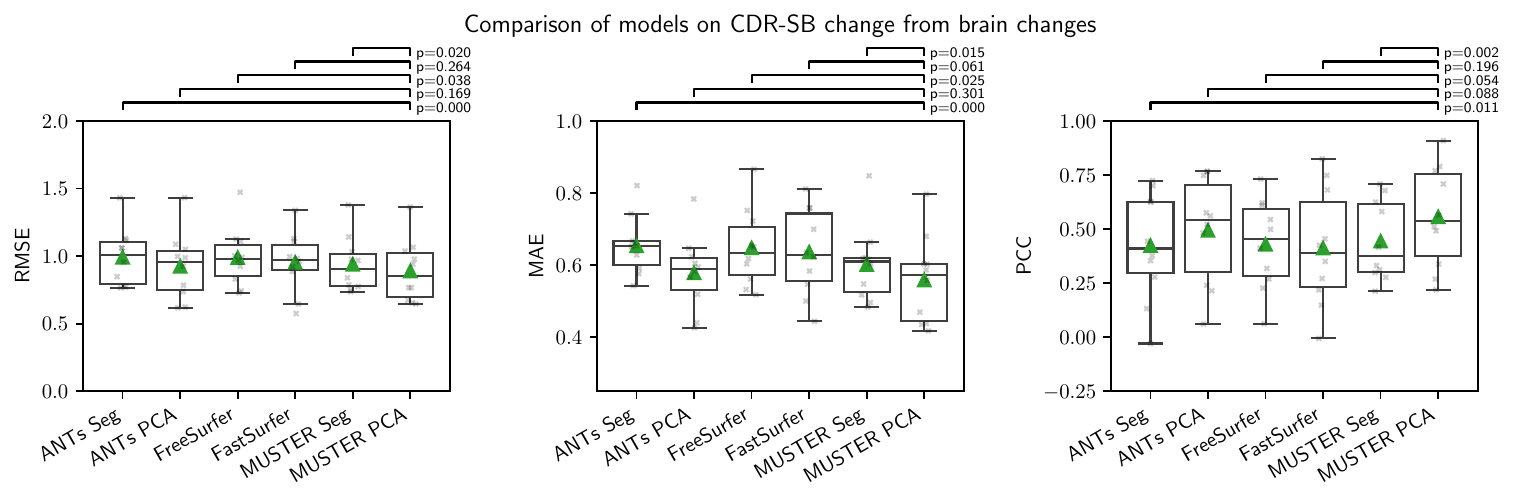}
    \caption{Performance metrics for the CDR-SB change models. P-value is shown between MUSTER PCA and the other methods using a paired t-test. RMSE: Root Mean Squared Error. Lower is better. MAE: Mean Absolute Error. Lower is better. PCC: Pearson's Correlation Coefficient}
    \label{fig:cdr_sb_change}
\end{figure}

\begin{figure}
    \centering
    \includegraphics[width=\linewidth]{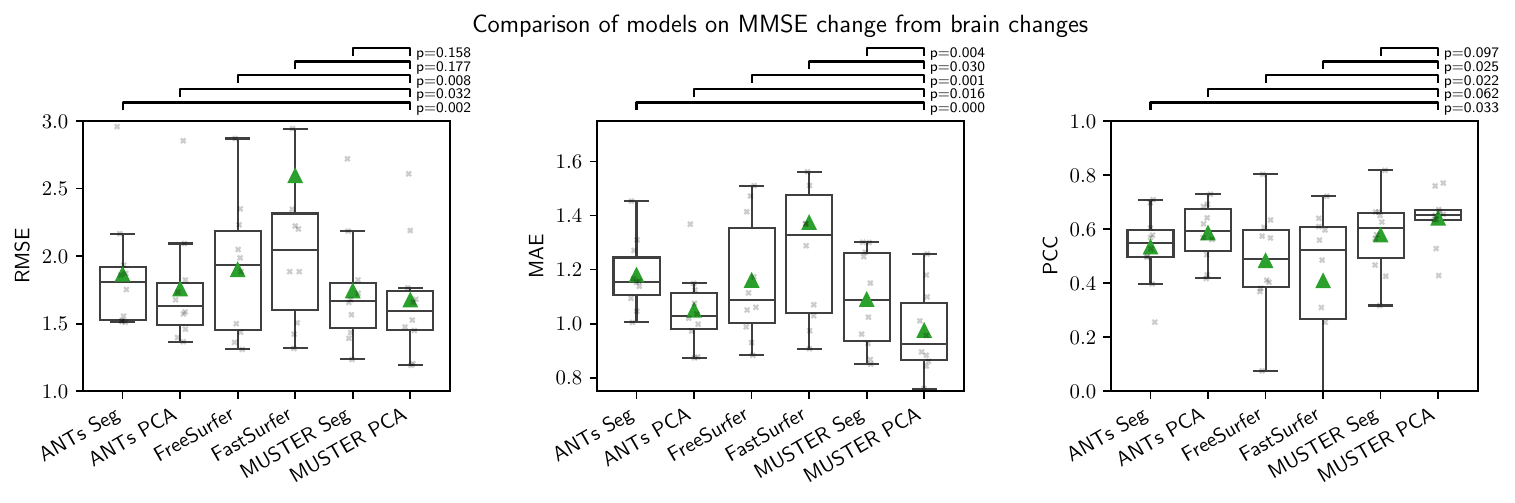}
    \caption{Performance metrics for the MMSE change models. P-value is shown between MUSTER PCA and the other methods using a related t-test. RMSE: Root Mean Squared Error. Lower is better. MAE: Mean Absolute Error. Lower is better. PCC: Pearson's Correlation Coefficient}
    \label{fig:mmse_change}
\end{figure}

\subsection{PCA of ADNI}
\begin{figure}
    \centering
    \includegraphics[width=1\linewidth]{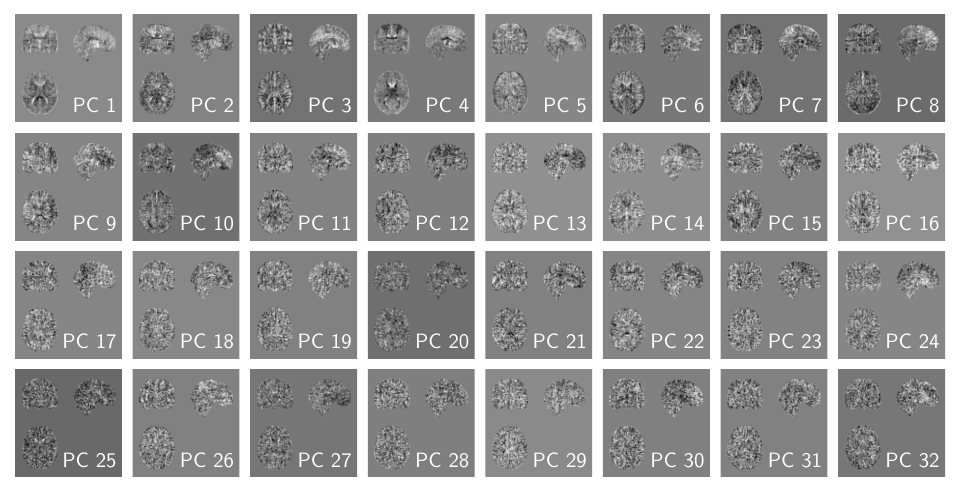}
    \caption{The 32 principle components of ADNI jacobian determinant using ANTs for longitudinal registration.}
    \label{fig:ants_pca}
\end{figure}

\begin{figure}
    \centering
    \includegraphics[width=1\linewidth]{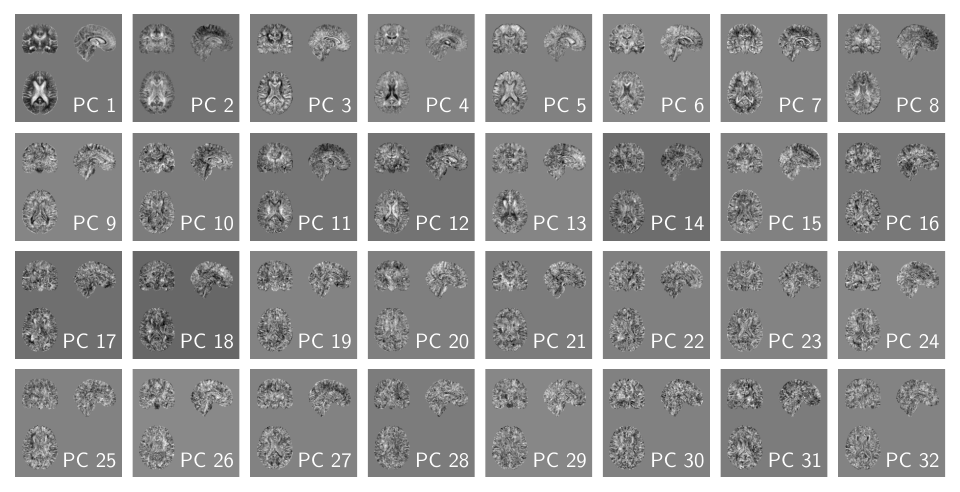}
    \caption{The 32 principle components of ADNI jacobian determinant using MUSTER for longitudinal registration.}
    \label{fig:muster_pca}
\end{figure}

\begin{figure}
    \centering
     \begin{subfigure}[b]{0.31\textwidth}
    \includegraphics[width=\linewidth]{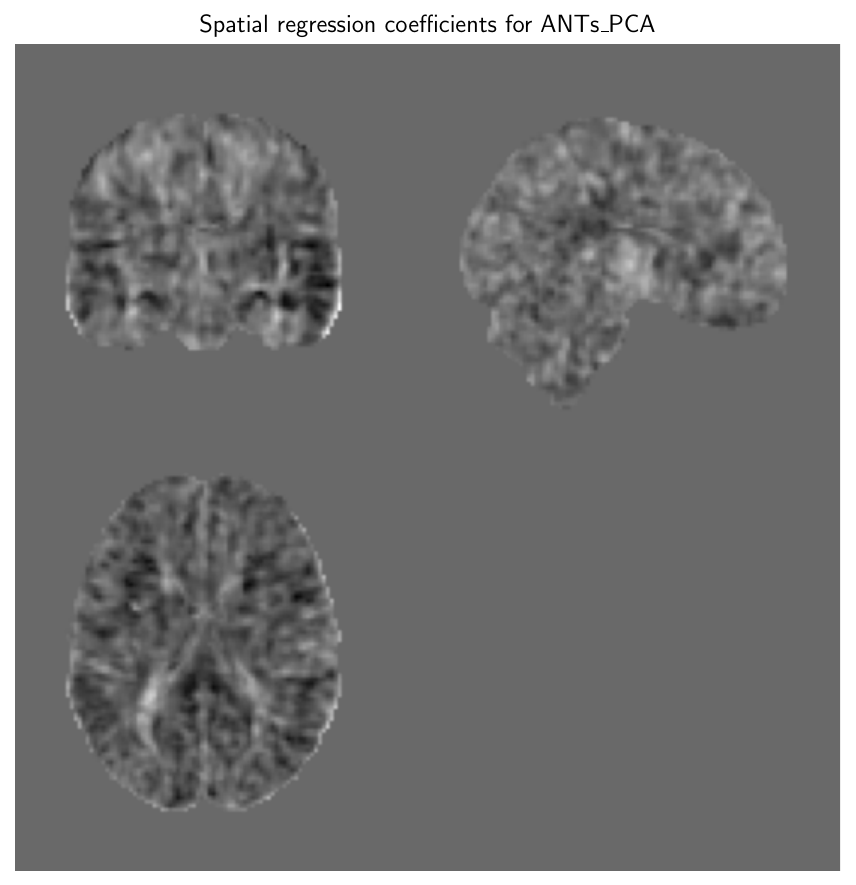}
    \end{subfigure}
    \begin{subfigure}[b]{0.31\textwidth}
    \includegraphics[width=\linewidth]{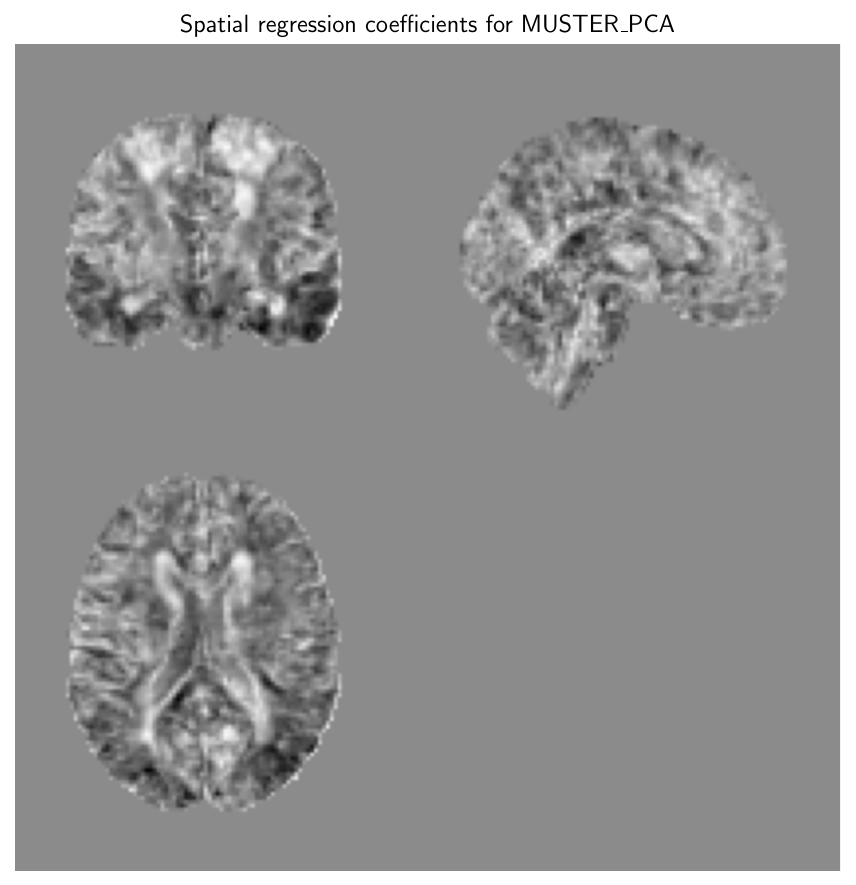}
    \end{subfigure}
    \caption{The spatial coefficients relating Jacobian determinants to cognitive change.}
    \label{fig:spatial_coeffs}
\end{figure}

\end{document}